\documentclass[journal ]{new-aiaa}
\usepackage[utf8]{inputenc}
\usepackage{textcomp}
\usepackage[version=4]{mhchem}
\usepackage{longtable,tabularx}
\usepackage{amsmath} 
\usepackage{amsxtra}     
\usepackage{amsfonts}
\usepackage{graphicx}    
\usepackage{siunitx}
\usepackage{rotating}
\usepackage{color}
\usepackage{epsfig}
\usepackage{verbatim}
\usepackage{natbib}      
\usepackage{acronym} 
\usepackage{setspace}    

\usepackage{titlesec}
\usepackage{verbatim}
\usepackage{booktabs}
\usepackage{setspace}
\usepackage{bigstrut}
\usepackage{lineno}
\usepackage{afterpage}
\usepackage{graphicx}
\usepackage{array}
\usepackage{multirow}
\usepackage{epsfig}
\usepackage{rotating}
\usepackage{algorithmicx}
\usepackage{color}
\usepackage[normalem]{ulem}
\usepackage{pstricks,pst-node}
\usepackage{multirow}
\usepackage{flushend}
\usepackage{amsbsy,epsfig}
\usepackage{lscape}
\usepackage[noend]{algpseudocode}
\usepackage{epstopdf}
\usepackage{mathtools}
\usepackage{framed}
\usepackage{algorithm}
\usepackage{hyperref}
\usepackage{multicol}
\usepackage{enumitem}
\usepackage{bm}
\usepackage{booktabs}
\usepackage{soul}
\usepackage{tabularx}
\usepackage{multicol}
\usepackage{placeins}
\usepackage{nomencl}
\usepackage{textcomp}
\usepackage{multicol}
\usepackage{siunitx} 
\usepackage{pdfcomment}
\usepackage{ar}
\usepackage{extramarks}
\usepackage{subcaption}
\usepackage{mwe}

\newcommand{\astar}[1]{$\text{A}^*_{#1}$}
\newcommand{\bitstar}[1]{$\text{BIT}^*_{#1}$}

\newcommand{\prmstar}{$\text{PRM}^*$}
\newcommand{\rrtstar}{$\text{RRT}^*$}
\newcommand{\fmtstar}{$\text{FMT}^*$}
\newcommand{\ptp}{$\text{PTP}$}

\definecolor{tab10-blue}{HTML}{1f77b4}
\definecolor{tab10-orange}{HTML}{ff7f0e}
\definecolor{tab10-green}{HTML}{2ca02c}
\definecolor{tab10-red}{HTML}{d62728}
\definecolor{tab10-purple}{HTML}{9467bd}
\definecolor{tab10-brown}{HTML}{8c564b}
\definecolor{tab10-pink}{HTML}{e377c2}
\definecolor{tab10-gray}{HTML}{7f7f7f}
\definecolor{tab10-olive}{HTML}{bcbd22}
\definecolor{tab10-cyan}{HTML}{17becf}
\makenomenclature
\title{Urban Metric Maps for Small Unmanned Aircraft Systems Motion Planning}

\author{Cosme A. Ochoa\footnote{PhD Candidate, Robotics Institute, Student Member {\tt\small cosme@umich.edu}} and Ella M. Atkins \footnote{Professor, Department of Aerospace Engineering, Fellow {\tt\small ematkins@umich.edu}}}

\affil{University of Michigan, Ann Arbor, MI, 48109, USA}

\begin{document}

\maketitle

\begin{abstract}
Low-altitude urban flight planning for small Unmanned Aircraft Systems (UAS) requires accurate vehicle, environment maps, and risk models to assure flight plans consider the urban landscape as well as airspace constraints. This paper presents a suite of motion planning metrics designed for small UAS urban flight. We define map-based and path-based metrics to holistically characterize motion plan quality. Proposed metrics are examined in the context of representative geometric, graph-based, and sampling-based motion planners applied to a multicopter small UAS. A novel multi-objective heuristic is proposed and applied for graph-based and sampling motion planners at four urban UAS flight altitude layers. Monte Carlo case studies in a New York City urban environment illustrate metric map properties and planner performance.  Motion plans are evaluated as a function of planning algorithm, location, range, and flight altitude.
\end{abstract}

\printnomenclature[3cm]

\nomenclature{$x, y, z$}{                           Aircraft position (inertial frame)  }
\nomenclature{$\mathcal{D}$}{                       Date-time information  }
\nomenclature{$\mathcal{L}$}{                       Total bounding box  }
\nomenclature{$\mathcal{W}$}{                       Weighting vector  }
\nomenclature{$\mathcal{H}_{\delta_r, \delta_z}$}{  Metric cost map set }
\nomenclature{$\mathcal{H}_{total}$}{               Total cost map  }
\nomenclature{$\mathcal{H}_{norm}$}{                Min-max normalized map  }
\nomenclature{$\delta_{b}$}{                        Bounding box buffer distance }
\nomenclature{$\delta_{c}$}{                        Graph connectivity   }
\nomenclature{$\delta_{d}$}{                        Time of day   }
\nomenclature{$\delta_{r}$}{                        Map resolution  }
\nomenclature{$\delta_{z}$}{                        Planning altitude  }
\nomenclature{$m_m, m_p$}{                          Map/path-based metrics  }
\nomenclature{$m_{gps}, c_{gps}, w_{gps}$}{         GPS pseudorange uncertainty metric/cost/weight  }
\nomenclature{$m_{lidar}, c_{lidar}, w_{lidar}$}{   Lidar-based visibility metric/cost  }
\nomenclature{$m_{obs}, c_{obs}$}{                  Obstacle occupancy metric/cost/weight  }
\nomenclature{$m_{pop}, c_{pop}, w_{pop}$}{         Overflown population density metric/cost  }
\nomenclature{$m_{risk}, c_{risk}, w_{risk}$}{      Obstacle proximity metric/cost/weight  }
\nomenclature{$m_{dist}, c_{dist}, w_{weight}$}{    Distance traveled metric/cost/weight  }
\nomenclature{$c$}{                                 Speed of light  }
\nomenclature{$\rho_{rc}$}{                         GPS receiver range  }
\nomenclature{$t_{rc}, t_{sat}$}{                   GPS receiver/satellite clock  }
\nomenclature{$\epsilon_{UERE}$}{                   GPS error term  }
\nomenclature{$x_{rc}$, $y_{rc}$, $z_{rc}$}{        GPS receiver position (inertial frame)  }
\nomenclature{$x_{sat}$, $y_{sat}$, $z_{sat}$}{     GPS satellite position (inertial frame)  }
\nomenclature{$N_{sats}$}{                          number of visible satellites  }
\nomenclature{$G_{gps}$}{                           GPS pseudorange linear system  }
\nomenclature{$X_{gps}, \Sigma_{gps}$}{ GPS pseudorange state vector/covariance  }
\nomenclature{$GDOP_{thresh}$}{                     GDOP threshold  }
\nomenclature{$b_{o}$}{                   Lidar beams' origin  }  
\nomenclature{$b_{lidar}$}{                         number of Lidar beams  }
\nomenclature{$k_{lidar}, s_{lidar}$}{              Number of Lidar scan positions/returns  }
\nomenclature{$\Omega_{obs}$}{                    Obstacle set  }
\nomenclature{$r_{lidar}$}{                         Lidar visible range  }
\nomenclature{$\beta$}{                             Lidar beam elevation angle  }
\nomenclature{$\mathcal{C}_{free}, \mathcal{C}_{obs}, \mathcal{C}_{tot}$}{              Obstacle-free/obstacle/total configuration space  }
\nomenclature{$pop_{census}$}{                      Census raw population count  }
\nomenclature{$pop_{mod}, pop_{norm}$}{                      Modified/maximum population count  }
\nomenclature{$\Gamma_{comm}, \Gamma_{resi}$}{      Commercial/residential area population modifier  }
\nomenclature{$\mathcal{B}, \hat{\mathcal{B}}$}{    Regular/buffered operating bounding box  }
\nomenclature{$d_{close}$}{                         Distance to closest obstacle surface }
\nomenclature{$d_{thresh}$}{                        Proximity risk distance threshold  }
\nomenclature{$\boldsymbol{\zeta}$}{                Flight plan/path  }
\nomenclature{$\mathcal{H}_{obs}$}{                 Obstacle occupancy map  }
\nomenclature{$\mathcal{H}_{gps}$}{                 GPS uncertainty map  }
\nomenclature{$\mathcal{H}_{lidar}$}{               Lidar visibility map  }
\nomenclature{$\mathcal{H}_{pop}$}{                 Population density map  }
\nomenclature{$\mathcal{H}_{risk}$}{                Proximity risk map  }
\nomenclature{$\mathcal{Q}_S, \mathcal{Q}_G$}{      Start/goal vehicle state  }
\nomenclature{$idx, idy, idz$}{                     State indices  }
\nomenclature{$d_{euc}, d_{oct}$}{                  Euclidean/octile distance  }
\nomenclature{$f, g, h$}{                           Total cost, cost-so-far, cost-to-go heuristic }
\nomenclature{$h_{dist}, h_{plus}$}{                Distance and multi-objective heuristics  }
\nomenclature{$\mathcal{G}, T, V, E$}{                    Search graph/tree/nodes/edges  }

\section{Introduction}

A motion planner constructs a feasible and efficient kinodynamic path through a potentially complex environment connecting an initial location to a target or goal state \cite{lavalle_planning_2006}. Motion planning algorithms have been used for a wide range of Unmanned Aircraft Systems (UAS) applications including search \& rescue \cite{colas_3d_2013, berger_innovative_2015}, reconnaissance missions \cite{obermeyer_path_2009, obermeyer_samplingbased_2012}, sense \& avoid \cite{reif_motion_1985, radmanesh_overview_2018}, and navigation through unmapped or uncertain environments \cite{rathbun_evolution_2002, butenko_cooperative_2003, dadkhah_survey_2012}. Motion planners can complement onboard sensor suites to aid in conflict resolution \cite{peinecke_deconflicting_2017} and alternative fail-safe protocols \cite{ochoa_failsafe_2017} for urban flight. Baseline flight plans are computed and approved prior to flight, but real-time planning may be required to effectively respond to changes in the mission, environment, and/or vehicle performance (e.g., system degradation, failure).
Motion planners typically optimize solutions over path distance, time, and obstacle/terrain avoidance with benchmarks as discussed in \cite{moll_benchmarking_2015}.  Recent papers have presented flight risk metrics that augment traditional distance/time/obstacle avoidance cost terms \cite{shan_samplingbased_2015,didonato_evaluating_2017, castagno_comprehensive_2018, ippolito2019dynamic,rudnick2019modeling}.

This paper proposes a suite of complementary motion planning metrics designed for urban multicopter flight that further augments distance/time and risk-based metrics. We define map-based ($m_{m}$) and path-based ($m_{p}$) metrics to generate holistic cost-minimum plans in representative geometric, graph-based, and sampling-based motion planners. Map-based metrics ($m_m$) describe the UAS operating environment by constructing a collection of GPS/lidar navigation performance, population density, and obstacle risk exposure maps. Traditional path-based metrics ($m_p$) account for UAS energy consumption and distance traveled along a planned path. This paper presents a detailed analysis of map-based and path-based metrics in Monte Carlo case studies.

Map-based metrics are derived offline from open-source geospatial, satellite imagery, and census data. Each database is processed and transformed into discretized metric maps representative of the borough of Manhattan in New York City at different map resolutions and small UAS (sUAS) above ground level (AGL) flight altitudes.  GPS satellite availability, lidar visibility, and risk to an overflown population are captured. 
Motion planning metric maps are examined with respect to a portfolio of motion planners. Distance-only and weighted multi-objective cost function results are compared.  To improve performance, a multi-objective heuristic function for graph-based and sampling-based path planners is proposed. Monte Carlo case study results are presented as a function of metric weightings, planner type, and urban canyon settings. Planner metric usage and solution path properties are discussed.

The contributions of this work are as follows:
\begin{itemize}[noitemsep]
    \item This paper defines a comprehensive suite of urban UAS flight planning metrics and describes how to transform open-source data into metric maps applicable across different motion planners. 
    \item This paper presents representative geometric, graph-based, and sampling-based motion planners and describes how metric maps are deployed in each. 
    \item A novel multi-objective heuristic function is defined to improve upon a traditional distance-only heuristic.  This heuristic is applied and evaluated in graph-based and sampling-based motion planners.
    \item Monte Carlo simulations are evaluated to analyze the properties of motion plans generated with different cost metrics and different planning algorithms.
\end{itemize}

Below, Sec. \ref{sec:relatedwork} summarizes related work followed by a problem statement (Sec. \ref{sec:problem-statement}). Sec. \ref{sec:metric-definitions} defines map-based and path-based motion planning metrics followed by a description of the process by which discretized feature maps are generated (Sec. \ref{sec:map-generation}). A representative portfolio of motion planners is defined in Sec. \ref{sec:planning-algorithms}, and our novel multi-objective admissible heuristic is introduced. Map-based metric results are presented in Sec. \ref{sec:results}.  Monte Carlo simulation process is summarized in Sec. \ref{sec:sim}, and path planning results are evaluated in Sec. \ref{sec:path-results}.   Sec. \ref{sec:conclusions} concludes the paper.

\section{Related Work}
\label{sec:relatedwork}
This section first discusses background in metrics relevant to small UAS urban motion planning followed by background in motion planning approaches to sUAS operating in and over urban environments.

\subsection{Planning Metrics}

Qualitative and quantitative metrics inform a planner about the vehicle, its environment, preferences and constraints. Algorithm metrics can be defined from learned performance models \cite{roberts_learned_2007, roberts_learning_2009}, statistical measures \cite{saxena_evaluating_2009}, abstract features \cite{jain_feature_1997}, and classical algorithm properties \cite{russell_artificial_2010} as summarized in Table \ref{tab:planning_properties}. Additional metrics can be defined to incorporate application-specific considerations.  

\begin{table}[h]
    \caption{Classical motion planning algorithm properties.}
    \label{tab:planning_properties}
	\centering
  	\begin{tabular}{ p{0.17\textwidth}  p{0.75\textwidth}  }
  	\toprule
        Property & Description  \\ \midrule
        Completeness & A solution is returned if one exists; otherwise, failure is returned.  \\
        Soundness & If a solution is returned, it is feasible.  \\
        Complexity & Memory usage and/or execution time measured with theoretical upper bounds and/or large-scale Monte Carlo simulation.  \\
        Kinodynamics  & Planning solutions are consistent with vehicular performance constraints.  \\
       Environment &   Description of environment as static or dynamic.  \\
        Uncertainty & Planner accounts for uncertainty in vehicle or environment states.  \\
        Optimality & A best solution is returned with respect to a given metric or combination of metrics.   \\
        \bottomrule
  	\end{tabular}
	
\end{table}

In practice, a motion planner should be complete and return an optimal feasible solution in real-time, if necessary, while satisfying all kinodynamic constraints. Motion planners trade off different objectives to find a balanced solution \cite{lunenburg_motion_2016}. Distance traveled and flight risks per Table \ref{tab:risk_types} may be considered. Distance traveled captures expected energy expenditure and estimated time of arrival (ETA) at a destination, while risk metrics may account for non-ideal vehicle and environment properties. Our work primarily considers an environment risk metric map since vehicle performance and weather are dynamic entities that do not map to fixed Earth-based coordinates. 
 
\begin{table*}[ht]
    \caption{Common risks encountered by small UAS.}
	\label{tab:risk_types}
	\centering
	\begin{tabular}{p{0.12\textwidth}p{0.42\textwidth}p{0.36\textwidth}}
		\toprule
		Type & Description & Examples \\
		\midrule
		System      & A hardware or software failure resulting in a system freeze, coding error, reboot, or component failure.
		& Deadlock \cite{gligor_deadlock_1980, singhal_deadlock_1989}, overheating \cite{dadvar_potential_2005}, electrical shorts \cite{mohla_electrical_1999}, software risks \cite{boehm_software_1991} \\
		
		Actuators   & Control surfaces are irresponsive or fail to reach a target configuration given a threshold.
		& Shaft failures \cite{bonnett_root_2000}, PWM relay errors \cite{richardeau_failurestolerance_2002}, pneumatic/hydraulic faults \cite{graves_spectral_2018} \\
		
		Sensors     & Onboard sensing tools provide inaccurate representations of the world around them.
		& Faulty sensors, obstructed view, drifting sensor readings, urban canyon effects  \\
		
		Weather     & Hazardous climate conditions influencing system sensing and/or performance.
		& Cold impact on batteries \cite{jaguemont_thermal_2016}, poor visibility, snow/ice, turbulent winds  \cite{watkins_gusts_2019}  \\
		Environment & Operating in hazardous areas that could potentially injure or harm
		nearby structures or people.
		& Proximity to buildings \cite{ancel_realtime_2017}, flying over people \cite{didonato_evaluating_2017}, navigating unmapped areas \\
		\bottomrule
	\end{tabular}
\end{table*}

For real-time aerospace applications, completeness, soundness, and bounded computational complexity are desired algorithm properties. Fixed-wing aircraft typically optimize cruise altitude (atmospheric density), airspeed, climb rate, lift/drag ratio \cite{puranik_aircraft_2020,mcclamroch_steady_2011}, and hazardous weather avoidance \cite{balachandran_flight_2016} but do not consider ground-based obstacles due to their substantial cruise altitude. Multicopter UAS operate at much lower altitudes thus typically optimize motions over clearance from obstacles, distance / time, and mission requirements \cite{tenharmsel_emergency_2017}. Communication \cite{mardani_communicationaware_2019} and navigation \cite{bopardikar_multiobjective_2014} metrics are key considerations where line-of-sight signals may be blocked. A Pareto front analysis offers insight into balancing competing metrics \cite{castagno_comprehensive_2018, mitchell_continuous_2003, shashimittal_threedimensional_2007, guigue_pareto_2010}.

\subsection{Motion Planning}

The following paragraphs summarize different motion planning strategies and their respective advantages and disadvantages for small UAS urban motion planning.


\emph{Geometric} motion planners provide rapid analytical solutions by constructing paths using points, lines, and arcs. In a two-dimensional Euclidean space, visibility graphs \cite{nilsson_mobius_1969, lozano-perez_algorithm_1979} can be used to generate minimum length paths from intersecting lines for a holonomic system. Dubins \cite{dubins_curves_1957} and Reeds-Shepp \cite{reeds_optimal_1990} curves account for nonholonomic turning constraints by adding turning radius arc segments to a path as needed. Geometric planners generate solutions rapidly but make simplifying assumptions, e.g., obstacle-free environments. 

\emph{Graph-based} planners search for solutions in a graph defined to assure mapped obstacle avoidance. A motion planning space can be covered with a uniform or nonuniform grid or with a roadmap, e.g., visibility graph \cite{lavalle_planning_2006}.  By connecting the start and goal configurations to the graph, the motion planning problem is reduced to searching the graph for a minimum-cost path. A* \cite{hart_formal_1968} and its variants (Dijkstra \cite{dijkstra_note_1959}, LPA* \cite{koenig_lifelong_2004}, ARA* \cite{likhachev_ara_2003}, D* Lite \cite{koenig_fast_2005}, Field D* \cite{ferguson_field_2007}, Theta* \cite{daniel_theta_2010}) are among the popular graph search strategies adapted to motion planning. Graph-based planners thrive in low-dimensional configuration spaces to provide optimal solutions with arbitrarily-complex cost functions and constraints implicitly handled in the graph. However, their performance advantage diminishes as the dimensionality of the motion planning state-space increases.

\emph{Sampling-based} planners use randomly drawn node samples from an underlying probability distribution to generate a local graph iteratively. Probabilistic roadmaps (PRMs) \cite{kavraki_probabilistic_1996} and rapidly exploring random trees (RRTs) \cite{lavalle_rapidlyexploring_1998, kuffner_rrtconnect_2000} paved the way for sampling algorithms aimed at managing the high dimensionality problem of graph-based planners.  Innovations in the past decade have resulted in asymptotically optimal variants  (e.g., \prmstar{}, \rrtstar{} \cite{karaman_samplingbased_2011}) with improved convergence rates demonstrated in \fmtstar{} \cite{janson_fast_2015} and \bitstar{} \cite{gammell_batch_2015}. These algorithms are probabilistically complete but may not offer solutions in the presence of narrow passages or dense obstacle sets.

\emph{Optimization-based} planning methods construct a solution by minimizing a cost function while satisfying constraints, i.e., the boundary value problem \cite{kiguradze_boundaryvalue_1988}. Potential field methods \cite{barraquand_numerical_1991, ge_dynamic_2002} ignore dynamics and optimize a distance-based gradient along competing goal-attractive and obstacle-repulsive manifolds. Optimal control \cite{spindler_motion_2002} applies physics-based constraints and costs to minimize time, energy, and potentially obstacle avoidance using smooth spatiotemporal mathematical functions. The functional nature of optimization-based methods supports analyzing nonlinear, multiple input-output, and time-varying systems but at the cost of computational complexity and convergence challenges. Model predictive control \cite{wang_fast_2010, howard_modelpredictive_2014, liu_path_2017} variants limiting computations to a finite future horizon and can use lookup tables to cache complex solutions for online use. Optimization methods are susceptible to the local minima; they are not guaranteed to converge to a satisficing or globally-optimal solution particularly in complex environments.

\section{Problem Statement}
\label{sec:problem-statement}

\begin{figure}[t]
    \centering
    \includegraphics[width = 0.90\textwidth]{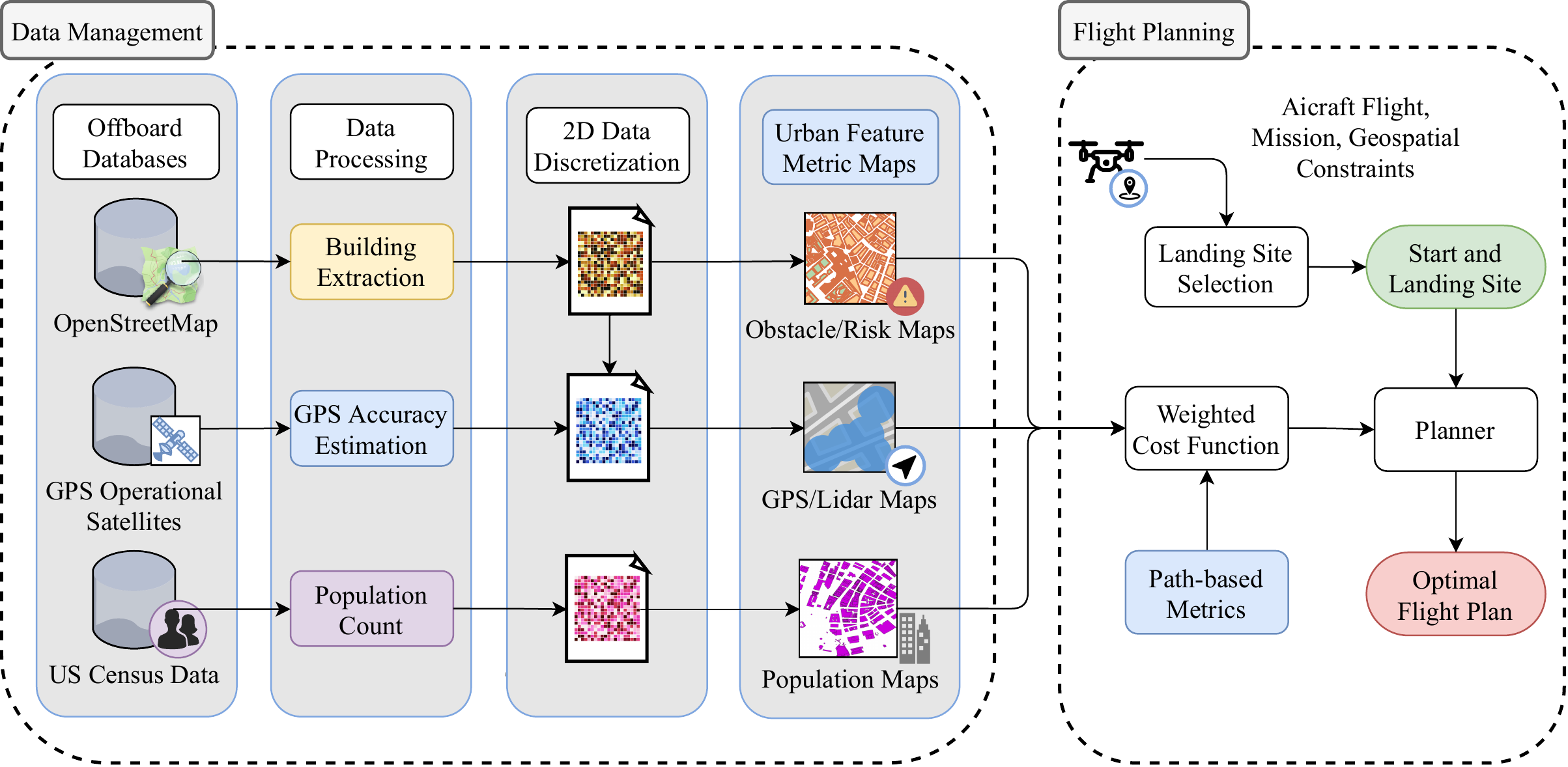}
    \caption{Data flow for map-based metric generation in data-driven multicopter flight planning.}
    \label{fig:data_planner_flow}
\end{figure}

This paper defines a suite of map-based metrics $m_{m}$ and path-based metrics $m_{p}$ to offer comprehensive environment and path cost for sUAS flight planners per Table \ref{tab:metric_types}.  Map-based metrics must be generated from a hybrid suite of data sources representing obstacles, sensor availability, and risk sources. Data must be processed, discretized, and converted into feature-rich metric maps, combined with  path-based metrics, to compute an optimal obstacle-free path to a targeted landing site as shown in Fig. \ref{fig:data_planner_flow}.

\begin{table}[h]
	\centering
	\caption{Motion planning metrics classified by type.}
	\begin{tabular}{c l c}
		\toprule
		Metric       & Description                          & Type     \\
		\midrule
		$m_{gps}$    & GPS pseudorange position uncertainty & map      \\
		$m_{lidar}$  & Lidar-based local map uncertainty    & map      \\
		$m_{obs}$    & Obstacle occupancy & map \\
		$m_{pop}$    & Overflown population estimate        & map      \\
		$m_{risk}$   & Proximity to obstacles en route      & map     \\
		$m_{dist}$   & Distance traveled along a path         & path     \\
		\bottomrule
	\end{tabular}
	\label{tab:metric_types}
\end{table}

For a given operating bounding box $\mathcal{L}$, a collection of metric maps $\mathcal{H}$ must be generated to describe all $m_m$ in Table \ref{tab:metric_types}. Each $\mathcal{H}$ is generated by explicitly calculating that metric value at every point characterized by a Cartesian grid over $\mathcal{L}$ with resolution $\delta_{r}$ for fixed flight altitude $\delta_{z}$. To explore these cost metrics, representative \emph{geometric}, \emph{graph-based}, and \emph{sampling-based} motion planning algorithms are defined for urban multicopter flight. Because a traditional Euclidean distance motion planning heuristic $h_{dist}$ does not provide information about any of the map-based metrics, a novel multi-objective heuristic $h_{plus}$ is proposed and compared to $h_{dist}$. A suite of Monte Carlo case studies in Manhattan illustrate metric map and motion plan properties in a representative urban environment at four UAS flight altitude layers. Motion plans are evaluated as a function of planner, location, range, and flight altitude.

\section{Metric Definitions}
\label{sec:metric-definitions}

\subsection{GPS Uncertainty}

GPS receivers communicate with a global navigation satellite system (GNSS) to estimate their geographical location using trilateration. Given receiver/satellite pairs, a pseudorange measurement is estimated as \cite{enge_global_1994}:
\begin{equation}
	\hat{\rho}_{rc, sat} =  \rho_{rc} + c(t_{sat} - t_{rc}) + \epsilon
	\label{eq:pseudorange}
\end{equation}
where  $\rho_{rc}$ is receiver range, $c$ is the speed of light, $t_{sat}$ and $t_{rc}$ are the satellite/receiver clock readings, and $\epsilon_{UERE}$ captures any User Equivalent Range Errors (UEREs), e.g., atmospheric, clock, signal, and multipath errors.

Geometric dilution of precision ($GDOP$) describes error propagation from satellite geometry: dispersed satellites reduce uncertainty while clustered satellites increase it \cite{santerre_impact_1991}. $GDOP$ can be expressed as:
\begin{equation}
	GDOP(x,y,z,t) = \sqrt{PDOP(x,y,z)^2 + TDOP(t)^2}
	\label{eq:gdop-simple}
\end{equation}
where $PDOP$ and $TDOP$ are position/time dilutions of precision, respectively. DOP values between 1 to 20 \cite{azami_classification_2013} quantify GPS reliability as summarized in Table \ref{tab:dop_ratings}. 

\begin{table}[h]
	\centering
	\begin{tabular}{p{0.1\linewidth}p{0.12\linewidth}p{0.67\linewidth}}
		\toprule
		DOP    & Rating    & Description                                           \\
		\midrule
		$1$  & Ideal     & Highest precision possible.                           \\
		$1-4$  & Excellent & Measurements are considered accurate except           
		for the most sensitive applications.                                       \\
		$4-6$  & Good      & Represents the minimum acceptable loss in accuracy.   \\
		$6-8$  & Moderate  & May still be used but only recommended in obstacle    
		free environments.                                                         \\
		$8-20$ & Fair      & Readings should be dismissed or only serve to compute 
		a rough estimate.                                                          \\
		$>20$  & Poor      & Unreliable and should not be used.                    \\
		\bottomrule
	\end{tabular}
	\caption{DOP Value Rating \cite{azami_classification_2013}.}
	\label{tab:dop_ratings}
\end{table}

For $n$ visible satellites, pseudo ranges offer a fast approximation of $PDOP$. Applying a first-order Taylor expansion to the true range, pseudorange $\hat{\rho}_{rc, i}$ and range $r_i$ to the $i$th satellite are computed as:

\begin{align}
	\hat{\rho}_{rc, i} & = \frac{x_{rc} - x_{sat, i}}{r_i} x_{rc} + \frac{y_{rc} - y_{sat, i}}{r_i} y_{rc} +
	\frac{z_{rc} - z_{sat, i}}{r_i} z_{rc} + c(t_{sat, i} - t_{rc}) \\
	r_i &= \sqrt{(x_{rc} - x_{sat, i})^2 + (y_{rc} - y_{sat, i})^2 + (z_{rc} - z_{sat, i})^2} 
\end{align}
where $x_{rc}$, $y_{rc}$, $z_{rc}$, $t_{rc}$ and $x_{sat, i}$, $y_{sat, i}$, $z_{sat, i}$, $t_{sat, i}$ are the positions/clock readings of the receiver and $i$th satellite respectively. Assuming vehicle and receiver co-location, this information can be expressed as a linear system  $G_{gps}$ and state vector $X_{gps}$:
\begin{equation}
	G_{gps} =
	\begin{pmatrix}
		\frac{x - x_1}{r_1} & \frac{y - y_1}{r_1} & \frac{z - z_1}{r_1} & -1 \\
		\frac{x - x_2}{r_2} & \frac{y - y_2}{r_2} & \frac{z - z_2}{r_2} & -1 \\
		\vdots              & \vdots              & \vdots              & \vdots \\
		\frac{x - x_n}{r_n} & \frac{y - y_n}{r_n} & \frac{z - z_n}{r_n} & -1 \\
	\end{pmatrix}
	\qquad
	X_{gps} = 
	\begin{pmatrix}
	    x \\
	    y \\
	    z \\
	    c \cdot t \\
	\end{pmatrix}
\label{eq:pseudorange-matrix}
\end{equation}
with a best linear unbiased estimator (BEST), covariance $\Sigma_{gps}  = \left (G^T_{gps} G_{gps} \right )^{-1} $ and dilutions of precision defined per \cite{langley_dilution_1999}:
\begin{equation}
	PDOP = \sqrt{\Sigma_{11, gps}^2 + \Sigma_{22, gps}^2 + \Sigma_{33, gps}^2}, ~~ TDOP = \sqrt{\Sigma_{44, gps}^2}
\end{equation}

Accounting for visible satelllites, we define a motion planning GPS map-based uncertainty metric $m_{gps}$ or cost $c_{gps}$ as:
\begin{align}
	m_{gps}(x,y,z,t) &= \frac{GDOP_{thresh} - \min(GDOP(x,y,z,t), GDOP_{cut})}{GDOP_{thresh} - 1} \\
	c_{gps}(x,y,z) &= 1 - m_{gps}(x,y,z)
	\label{eq:gps_metric}
\end{align}
where $GDOP_{thresh}$ is a worst-case cutoff value for safe flight.

\subsection{Lidar Visibility}
Lidar provides a local obstacle point cloud to assure safe navigation through complex spaces and support local-area mapping. In GPS-denied areas, lidar \cite{rufa_unmanned_2016} can be used for inertial navigation by tracking mapped buildings and other landmarks. Lidar uses a laser's reflection time to estimate distances to objects. Lidar can be configured as a dome or cylindrical puck for local and longer-range sUAS applications.

The puck configuration modeled in this work uses $b_{lidar}$ equiangular beams that revolve to scan at $k_{lidar}$ equiangular positions capturing $n_{lidar} = b_{lidar} \cdot k_{lidar}$ points per revolution. Because $k_{lidar} >> 1$, $n_{lidar}$ is impractical for metric normalization, we propose number of returned scan readings (where an obstacle is within lidar range) as a lidar metric. A scan reading is recorded if any beam of the $j$th scan, $j = 1,2, \cdots, k_{lidar}$, intersects an obstacle in $\Omega_{obs}$ within range $r_{lidar}$ from the sUAS:

\begin{equation}
    scan(j) = 
    \begin{cases}
		1, & \text{if }  \exists i \text{ s.t. } \overleftrightarrow{b_{o}b_{i,j}} \cap \Omega_{obs} \neq \emptyset\\
		0, & \text{otherwise} 
	\end{cases}
\end{equation}{}
where $i \in \{1, 2, \ldots, b_{lidar} \}$, $b_{o}$ is the origin point of all beams, and $b_{i,j}$ is the $i$th lidar beam point for the $j$th scan a distance $r_{lidar}$ away with an elevation angle $\beta_{i}$.

A count of total scan returns $s_{lidar}(x,y,z,r_{lidar}) = \sum_{1}^{k_{lidar}} scan(j)$ is then compared to the total number of possible scan returns in the following lidar metric $m_{lidar}$ or cost $c_{lidar}$:
\begin{align}
	m_{lidar}(x,y,z, r_{lidar}) &= \frac{s_{lidar}(x,y,z, r_{lidar})}{k_{lidar}} \\ c_{lidar}(x,y,z, r_{lidar}) &= 1 - m_{lidar}(x,y,z,r_{lidar})
\end{align}

\subsection{Obstacle Occupancy}
Obstacle maps allow motion planners to define free $\mathcal{C}_{free}$ and obstacle $\mathcal{C}_{obs}$ configuration spaces. We define an obstacle occupancy metric $m_{obs}$ to penalize flight paths with points that intersect obstacles such that:
\begin{equation} 
	m_{obs}(x,y,z) = c_{obs}(x,y,z) = 
	\begin{cases}
		0, & \text{if } (x,y,z) \cap \mathcal{C}_{obs} = \emptyset \\
		1, & \text{otherwise} 
	\end{cases}
\end{equation}


\subsection{Population Density}
Flying low imposes a nontrivial risk to the overflown population. Population metric $m_{pop}$ estimates expected normalized population density for each weekday. Population can be estimated from government census data \cite{census_tiger_2010} or dynamic sources such as mobile phone activity \cite{didonato_evaluating_2017}. For Manhattan, turnstile and taxi data have also been used to estimate population \cite{fung_manhattan_2019}. Similar information is not available across multiple cities, so we propose extrapolating population estimates directly from census data.

\begin{table}[h]
    \caption{Dynamic population estimates in millions for Manhattan in 2010. \cite{moss_dynamic_2012}.}
    \label{tab:pop_day_night_table}
    \centering
    \begin{tabular}{c c c}
        \toprule
                & Work Week & Weekend \\ 
        \midrule
        Daytime & 3.94 & 2.90 \\
        Nighttime & 2.05 & 2.05 \\
        \bottomrule
    \end{tabular}
    
\end{table}

Population estimates for Manhattan are presented in Table \ref{tab:pop_day_night_table}.  A city's population varies throughout the day. Due to typical work hours, e.g., 9-to-5, population estimates in commercial areas are higher during the day. As people return home after work residential areas become densely populated during the evening. To estimate occupancy for each census map grid, we assume census data $pop_{census}$ for nighttime population and modify daytime population by a scaling factor $\Gamma$ determined based on area zoning (commercial $\Gamma_{comm}$ or residential $\Gamma_{resi}$) such that:
\begin{align}
    pop_{mod}(x,y, \delta_{d}, \Gamma) &= 
    \begin{cases}
        \Gamma \cdot pop_{census}[\kappa(x,y)] & \text{if } \delta_{d} = day \\
        pop_{census}[\kappa(x,y)]  & \text{if } \delta_{d} = night
    \end{cases}
    \label{eq:popgamma}
\end{align}
where $\delta_{d}$ denotes time of day and $\kappa(\cdot)$ is an indexing function relating census index to world coordinates.

The following population density metric and cost pair is then defined:
\begin{equation}
	m_{pop}(x,y, \delta_{d}, \Gamma) = \frac{pop_{mod}(x,y, \delta_{d}, \Gamma)}{pop_{norm}(\mathcal{B}, \delta_{d})}  \qquad c_{pop}(x,y) = m_{pop}
	\label{eq:metric_pop}
\end{equation}
where $pop_{norm}(\cdot)$ is maximum daytime or nighttime population density over bounding region $\mathcal{B}$.

\subsection{Risk Proximity Metric}
For this work risk is simply defined as proximity to nearby buildings or terrain with a threshold-based rectifier function. A building map is used to compute the distance to the closest obstacle surface, $d_{close}(x,y,z)$, for each map grid or point in space. For a specified distance threshold, $d_{thresh}$, a proximity risk is defined as:
\begin{equation}
	m_{risk}(x,y,z) = \min \left ( \frac{d_{close}}{d_{thresh}}, 1 \right ) \qquad c_{risk}(x,y,z) =  1 - m_{risk}(x,y,z)
	\label{eq:proximity_risk}
\end{equation}

\subsection{Distance-based Path Metric}
The expected distance traversed is given by:
\begin{equation}
	m_{dist}(t_0, t_f) = \int_{t_0}^{t_f} |v(t)| dt
\end{equation}
where $t_0$ and $t_f$ are initial and final planned flight times and $v(\cdot)$ is velocity magnitude. This function can also be written as a summation of $N$ segment lengths over planned flight path $\boldsymbol{\zeta}$:
\begin{equation}
	m_{dist}(\boldsymbol{\zeta}, N) = \sum_{i=1}^{N} \sqrt{ (\zeta_{x,i} - \zeta_{x, i-1})^2 + (\zeta_{y,i} - \zeta_{y, i-1})^2 + (\zeta_{z,i} - \zeta_{z, i-1})^2}
\end{equation}
where $\zeta_i=(\zeta_{x,i}, \zeta_{y,i}, \zeta_{z,i})$ is the $i$th point in path $\boldsymbol{\zeta}$.


\section{Map Generation}
\label{sec:map-generation}

\begin{figure}[h]
    \centering
    \includegraphics[trim=5cm 2.5cm 5cm 1cm, clip, height=0.3\textheight]{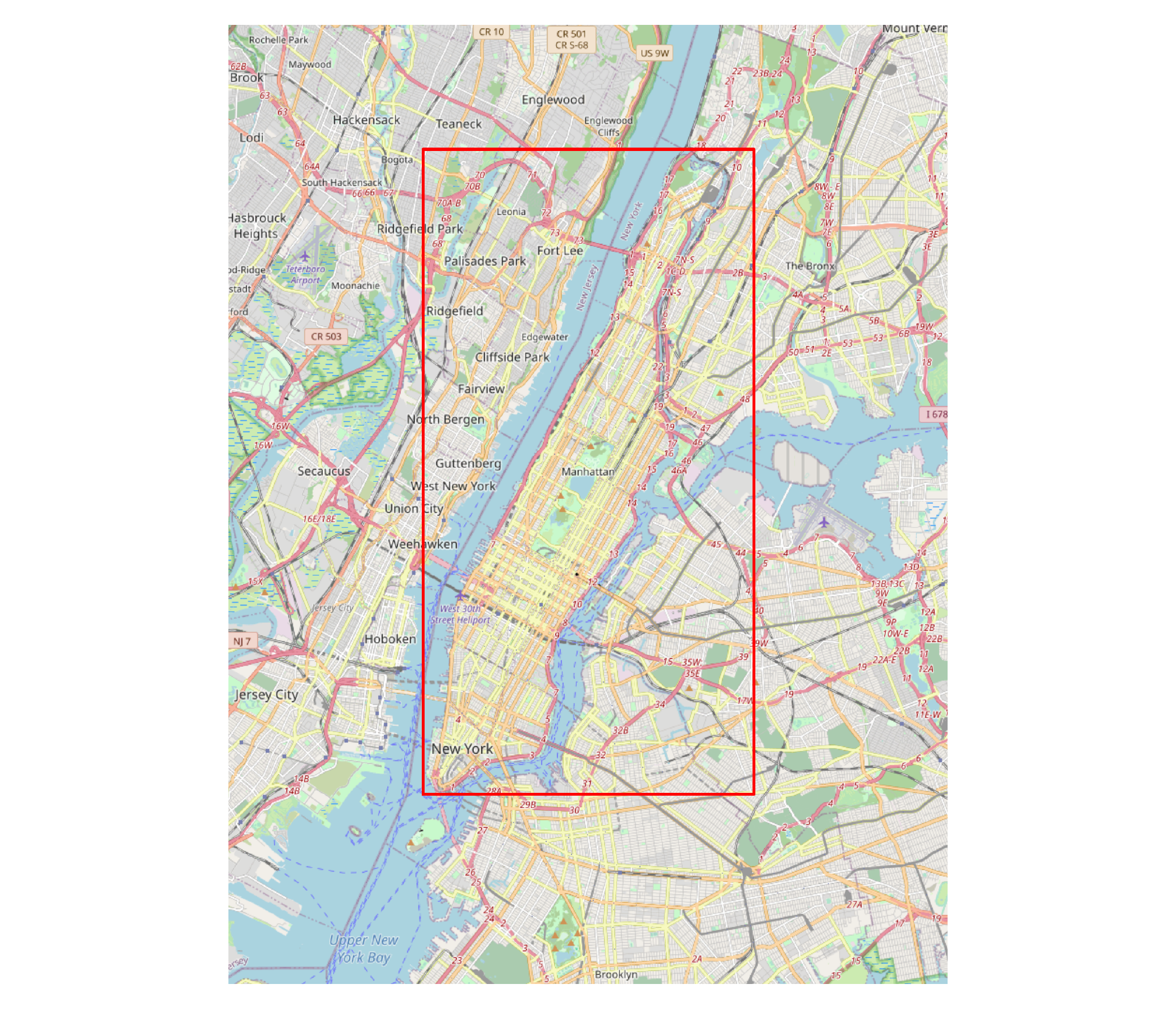}
    \caption{Planning configuration space area $\mathcal{L}$ for Manhattan case studies.}
    \label{fig:manhattan_bbox}
\end{figure}

Each Cartesian map of specified resolution defines a metric value for each spatial grid. For this investigation, metric maps cover an area $\mathcal{L}$ with a width 10km and height of 20km centered in Manhattan per Fig. \ref{fig:manhattan_bbox}.
Maps with 2m, 5m, and 10m resolution were generated. The 2m value coincides with current small UAS positioning and obstacle avoidance (trajectory tracking) accuracies. Height-dependent metrics were computed for UAS flight altitudes of 20m, 60m, 122m (current FAA maximum altitude for sUAS operations), and 600m AGL (above ground level), capturing low, medium, high, and ceiling-altitude flight. Note that cost map equivalents for each metric map can be computed by following the metric-to-cost conversions presented in the previous section.

\subsection{Obstacle Maps}
OpenStreetMap (OSM) \cite{haklay_openstreetmap_2008} data was processed to extract a building-based obstacle map $\mathcal{H}_{obs}$ from ways and relations using attribute labels. OSM data was converted to a local UTM 18N (EPSG:32618) coordinate reference system (CRS). The Universal Transverse Mercator (UTM) coordinate projection allows metric calculations directly defining axes (easting, northing) in meters. Extracted polygons $\Omega_{obs}$ were rasterized at each map resolution.  The height of the $k$th extracted polygon $z_k$ located at grid point $(x,y)$ was compared to UAS flight altitude $z$ such that:
\begin{equation}
    \mathcal{H}_{obs}(x,y,z) = 
    \begin{cases}
    1 & \text{if } z_k \geq z \\
    0 & \text{otherwise}
\end{cases}
\end{equation}

\subsection{GPS Maps}

GPS metric maps $\mathcal{H}_{gps}$ describe expected GPS accuracy for the Manhattan urban canyon. For a given grid point $(x,y,z)$ and date/time information $\mathcal{D}$, positions of overhead satellites are predicted using CelesTrak \cite{kelso_celestrak_1985} and Skyfield \cite{rhodes_skyfield_2020}. Rays are cast to above-horizon satellites and checked for collisions against extruded buildings in $\Omega_{obs}$. With less than four visible satellites ($N_{sats} < 4$), $m_{gps}$ is set to zero; otherwise the GPS pseudorange and covariance matrices are used to calculate $m_{gps}$:
\begin{equation}
    \mathcal{H}_{gps}(x,y,z) = 
    \begin{cases}
    m_{gps}(x, y, z) & \text{if } N_{sats} \geq 4 \\
    0 & \text{otherwise}
\end{cases}
\end{equation}

\subsection{Lidar Maps}

Lidar metric maps $\mathcal{H}_{lidar}$ estimate metric $m_{lidar}$, the expected percentage of lidar range returns. It is assumed that the vehicle is equipped with $b_{lidar}$ beams configured in a parallel configuration, i.e., the aircraft's $z_{body}$ and the lidar's rotation axis are parallel. Hence, the ratio of scan returns per revolution at each grid point $(x,y,z)$ is given by:

\begin{equation}
    \mathcal{H}_{lidar}(x,y,z, r_{lidar}) = m_{lidar}(x, y, z, r_{lidar})
\end{equation}

\subsection{Population Maps}
\label{sec:mm-pop-maps}

\begin{figure}[h]
    \centering
    \begin{subfigure}{0.45\textwidth}
        \centering
        \includegraphics[width=0.60\textwidth, height=0.30\textheight]{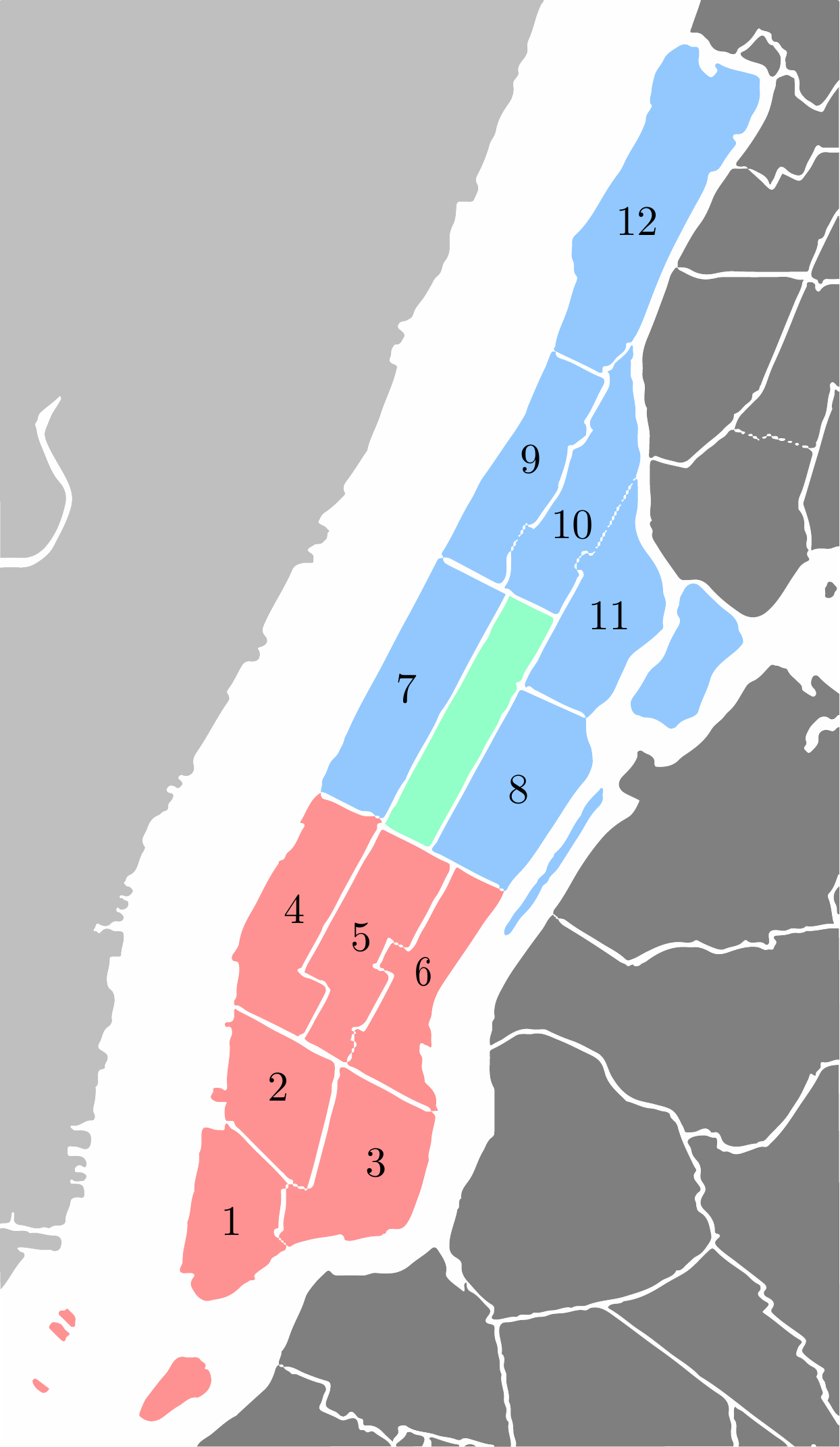}
        \caption{Figure of Manhattan district types.  Commercial regions are red; residential regions are blue.  Central Park with no permanent tenants is green.}
    \end{subfigure}
    \hfill
    \begin{subfigure}{0.45\textwidth}
        \centering
        \includegraphics[width=0.60\textwidth, height=0.30\textheight]{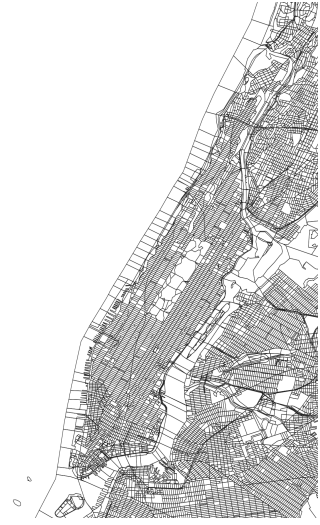}
        \caption{Manhattan census data reported in distinct polygonal regions.}
    \end{subfigure}
    \caption{Manhattan community districts and census data blocks.}
    \label{fig:manhattan-districts}
\end{figure}

Population metric maps $\mathcal{H}_{pop}$ are computed based on zoning and census data compiled into the normalized population metric $m_{pop}$. Census values are adjusted by $\Gamma$ as described in Eq. \ref{eq:popgamma} to adjust for commuting patterns between commercial $\Gamma_{comm}$ and residential $\Gamma_{resi}$  areas.
Manhattan is divided into twelve districts starting at its southernmost neighborhood, i.e., the Financial District, to its northernmost neighborhood, i.e., Harlem, as shown in Fig. \ref{fig:manhattan-districts}. The lower districts (1-6) are composed of businesses, government buildings, and tourist attractions. In contrast, the upper districts (7-12) consist mostly of single and multi-family residences. Defined by NYC Department of City Planning \cite{community_districts_2013},  the twelve districts are labeled as shown on Table \ref{tab:nyc_districts}.

\begin{table}[h]
    \caption{Manhattan districts with their primary zoning types.}
	\label{tab:nyc_districts}
	\centering
  	\begin{tabular}{l l l}
  	\toprule
  	Number & Neighborhoods & Type\\
  	\midrule
  	01 & Financial District, Civic Center & Commercial \\
  	02 & West Village, Greenwich Village, Soho & Commericial \\
  	03 & Chinatown, East Village, Noho & Commericial \\
    04 & Chelsea, Clinton, Hell's Kitchen & Commericial \\
    05 & Union Square, Madison Square, Times Square & Commericial \\
    06 & Gramercy, Murray Hill, Turtle Bay & Commericial \\
    07 & Lincoln Square, Upper West Side, Manhattan Valley & Residential \\
    08 & Lenox Hill, Upper East Side, Yorkville & Residential \\
    09 & Morningside Heights, Hamilton Heights & Residential \\
    10 & Central Harlem & Residential \\
    11 & East Harlem & Residential \\
    12 & Inwood, Washington Heights & Residential \\
  	\bottomrule
  	\end{tabular}
	
\end{table}

Population data for this study was derived from the 2010 United States Census \cite{census_tiger_2010}. The WGS84 CRS census block polygons represent the smallest geographic unit used by the US Census Bureau to estimate the number of residents in a block. Each census block entry includes a cumulative population count for that block and is assigned a district number 1-12 if the census block and district outline fully intersect. Any census block overlapping multiple outlines is assigned the district polygon's label with the largest intersection by area. Any census block within $\mathcal{L}$ but not in Manhattan, i.e., the Bronx or Queens, is given a district label of $13$ and labeled as residential. All geospatial data is converted to the UTM 18N CRS for consistency.  The population cost map is then defined as:

\begin{equation}
    \mathcal{H}_{pop, \delta_{d}}(x,y) = 
    \begin{cases}
        m_{pop}(x, y, \delta_{d}, \Gamma_{comm}) & \text{if } label[x,y] = \text{commercial} \\
        m_{pop}(x, y, \delta_{d}, \Gamma_{resi}) & \text{if } label[x,y] = \text{residential}
\end{cases}
\end{equation}

\subsection{Risk Maps}

The final metric map set $\mathcal{H}_{risk}$ quantifies building obstacle risks in the urban canyon as a function of the proximity risk metric $m_{risk}$ as shown below: 
\begin{equation}
    \mathcal{H}_{risk}(x,y,z) = m_{risk}(x, y, z)
\end{equation}

\subsection{Composite Metric Maps}
All the  metric maps described above are collected into set $\mathcal{H}_{\delta_r, \delta_z}$ defined by: 
\begin{equation}
    \mathcal{H}_{\delta_r, \delta_z} =
    \begin{pmatrix}
        \mathcal{H}_1 \\
        \mathcal{H}_2 \\
        \mathcal{H}_3 \\
        \mathcal{H}_4 \\
        \mathcal{H}_5 \\
    \end{pmatrix}_{\delta_r,\delta_z} =  
    \begin{pmatrix}
        \mathcal{H}_{obs} \\
        \mathcal{H}_{gps} \\
        \mathcal{H}_{lidar} \\
        \mathcal{H}_{pop, \delta_d} \\
        \mathcal{H}_{risk} \\
    \end{pmatrix}_{\delta_r, \delta_z}
    \label{eq:maps}
\end{equation}
where $\delta_r$ is map resolution and $\delta_z$ is UAS flight altitude assumed constant for each planning instance in this work.  A distinct  $\mathcal{H}_{\delta_r, \delta_z}$ is stored for each $(\delta_r, \delta_z)$ used in our case studies, and time of day $\delta_d$ as needed.

\section{Planning Algorithms}
\label{sec:planning-algorithms}
\subsection{Point-to-Point: \ptp{}}

The simplest path a multicopter can take is direct, i.e., point-to-point (\ptp{}). 
$\boldsymbol{\Lambda}_{\text{PTP}} = (\mathcal{L}, \mathcal{Q}_S, \mathcal{Q}_G, \mathcal{H}_{\delta_r, \delta_z}, \mathcal{W}, \delta_z, \delta_r)$ defines all relevant multicopter \ptp{} flight planning parameters where $\mathcal{W}$ is a cost weighting vector defined below.
\ptp{} is a simple geometric construct that assumes no obstacles are present.  A \ptp{} solution must therefore be post-processed to check for obstacle collisions and evaluate path cost.  The operating environment is described by the collection of metric maps $\mathcal{H}_{\delta_r, \delta_z}$ defined above. Each map is rasterized with metric values generated for each grid in the map search space  $\mathcal{L}$ at a given height and resolution pair $(\delta_z, \delta_r)$.
Using start and goal positions $\mathcal{Q}_S$ and $\mathcal{Q}_G$, the path's grid-based map indices $(\mathcal{Q}_{k,idx}, \mathcal{Q}_{k,idy})$ given origin $(\mathcal{L}_{x,min}, \mathcal{L}_{y,min})$ are calculated as:
\begin{align}
    \mathcal{Q}_{l,idx} &=  \left \lfloor \frac{\mathcal{Q}^{'}_{k,x}  - \mathcal{L}_{x,min}}{\delta_r} \right \rfloor \quad \mathcal{Q}^{'}_{l,x}(\alpha_{k,x}) =  \mathcal{Q}_{S,x} +   \alpha_{l,x} \\
    \mathcal{Q}_{l,idy} &=  \left \lfloor \frac{\mathcal{Q}^{'}_{k,y}  - \mathcal{L}_{y,min}}{\delta_r} \right \rfloor \quad \mathcal{Q}^{'}_{l,y}(\alpha_{k,y}) =  \mathcal{Q}_{S,y} +   \alpha_{l,y}
    \label{eq:index-conversion}
\end{align}
where $\alpha_{k,x}$ and $\alpha_{k,y}$ are component-wise steps from $\mathcal{Q}_{S}$ to $\mathcal{Q}_{G}$ for $l = 0, 1, \ldots \lceil \frac{\lambda}{\delta_r} \rceil $:
\begin{equation}
    \alpha_{l,x}  = \begin{cases}
\lambda \cos(\theta) & \text{if } l = \lceil \frac{\lambda}{\delta_r} \rceil\\
l \delta_r \cos(\theta)  & \text{otherwise}
\end{cases}  \qquad     \alpha_{k,x}  = \begin{cases}
\lambda \sin(\theta) & \text{if } l = \lceil \frac{\lambda}{\delta_r} \rceil\\
l \delta_r \sin(\theta)  & \text{otherwise}
\end{cases}
\end{equation}
where $\theta = \text{atan2}(\mathcal{Q}_{G,y} - \mathcal{Q}_{S,y}, \mathcal{Q}_{G,x} - \mathcal{Q}_{S,x})$ and $\lambda = \sqrt{(\mathcal{Q}_{G,y} - \mathcal{Q}_{S,y})^2 + (\mathcal{Q}_{G,y} - \mathcal{Q}_{S,y})^2} $.
Altitude $\delta_z$ is considered constant at one of the four designated layers for this study.

To test validity, a \ptp{} solution path $\zeta$ is masked onto obstacle map $\mathcal{H}_{obs} \in \mathcal{H}_{\delta_r, \delta_z}$. If any masked index has a non-zero value, i.e., $\mathcal{H}_{obs}(\mathcal{Q}_{l,idx}, \mathcal{Q}_{l,idy}, \mathcal{Q}_{l,idz}) > 0$, the path $\zeta$ is invalid; otherwise its cost is calculated. Total path cost $f(\zeta)$is defined by:
\begin{equation}
    f(\zeta) =  \sum_{i \in l} c(\mathcal{Q}_{i-1}, \mathcal{Q}_{i})
    \label{eq:straight-path-cost}
\end{equation}
where $c(\cdot)$ is the transition cost between adjacent states.  When using grid-based maps, the cost of moving between grids is described by the cost maps in $\mathcal{H}_{\delta_r, \delta_z}$. Given map indices $(idx, idy, idz)$ costs can be computed, weighted with vector $\mathcal{W}$, and summed.
The transition cost from $\mathcal{Q}_{i-1}$  to $\mathcal{Q}_i$ is then given by:
\begin{equation}
    c(\mathcal{Q}_{i-1}, \mathcal{Q}_i) = w_{0} d_{euc}(\mathcal{Q}_{i-1},\mathcal{Q}_i) + \sum_{j=1}^{k} w_{j}\mathcal{H}_j(\mathcal{Q}_{i,idx}, \mathcal{Q}_{i,idy}, \mathcal{Q}_{i,idz})
    \label{eq:c_cost_fun}
\end{equation}
where $d_{euc}(\cdot)$ is the Euclidean distance between states.  Per Eq. \ref{eq:maps}, $k=4$ cost metric maps for our planning case studies.

\subsection{Graph-based Planning: \astar{}}

\astar{} \cite{hart_formal_1968} is a discrete graph-based informed search algorithm popular for its completeness, optimality, and spatial efficiency. \astar{} searches a graph $\mathcal{G}$ to find a sequence of edge transitions that optimally navigates $\mathcal{G}$ from a start node $\mathcal{Q}_S$ to a goal node $\mathcal{Q}_G$. In motion planning, this sequence of edge transitions is equivalent to the desired path $\zeta$.  The \astar{} motion planning problem is defined by:
\begin{itemize}
    \item \emph{Parameters}: $\boldsymbol{\Lambda}_{\text{A}^*} = (\mathcal{L}, \mathcal{Q}_S, \mathcal{Q}_G, \mathcal{H}_{\delta_r, \delta_z}, \mathcal{W}, \delta_z, \delta_r, \delta_c)$
    \item \emph{Search Graph}: $\mathcal{G} = (V,E)$
    \item \emph{Total Cost Function}: $f(\cdot)= g(\cdot) + h(\cdot)$
\end{itemize}
where $\delta_c$ defines map cell adjacency for search graph $\mathcal{G}$, and $(V,E)$ are the nodes and edges forming $\mathcal{G}$, respectively.  $g(\cdot)$ is the cost function from the start node to the current search node, and $h(\cdot)$ is a heuristic function estimating cost from the current search node to the goal node.  Graph vertices $V$ are defined by discretizing $\mathcal{L}$ with resolution $\delta_r$. In an obstacle-free environment a maximum of $\frac{\mathcal{L}_{dx} \mathcal{L}_{dy}}{\delta_r^2}$ map grids may be traversed.  Configuration space $\mathcal{C}_{tot}$ is then:
\begin{equation}
    \mathcal{C}_{tot} = \left \{ \mathcal{Q}_l \; | \; \forall i, j \wedge \mathcal{Q}_{l,x} = i \frac{\mathcal{L}_{dx}}{\delta_r} \wedge \mathcal{Q}_{l,y} = j \frac{\mathcal{L}_{dy}}{\delta_r} \right \}
\end{equation}
where $i = 1, 2, \ldots, \mathcal{L}_{dx} / \delta_r$ and $j = 1, 2, \ldots,\mathcal{L}_{dy} / \delta_r$ such that $l = (i-1)\mathcal{L}_{dy} / \delta_r + j$.
Nodes with obstacle conflicts given by $\mathcal{C}_{obs} \subseteq \mathcal{C}_{tot}$ are defined as:
\begin{equation}
    \mathcal{C}_{obs} = \left \{ \mathcal{Q} \; | \; \mathcal{Q} \in \mathcal{C}_{tot} \wedge \mathcal{H}_{obs}(\mathcal{Q}_{idx}, \mathcal{Q}_{idy}, \mathcal{Q}_{idz}) = 1 \right \}
\end{equation}
All nodes with conflicts must be removed from the search-space; the obstacle-free configuration space $\mathcal{C}_{free}$ is then given by:
\begin{equation}
    V = \mathcal{C}_{tot} \setminus \mathcal{C}_{obs} \equiv \mathcal{C}_{free}
\end{equation}

\begin{figure}[h]
    \centering
    \begin{subfigure}[t]{.48\textwidth}
		\centering
    \includegraphics[width = 0.60\textwidth]{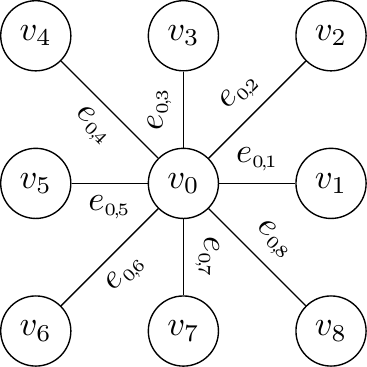}
        \caption{Fully connected graph.}
        \label{fig:full-graph}
    \end{subfigure}
    \hfill
    \begin{subfigure}[t]{.48\textwidth}
		\centering
    \includegraphics[width = 0.60\textwidth]{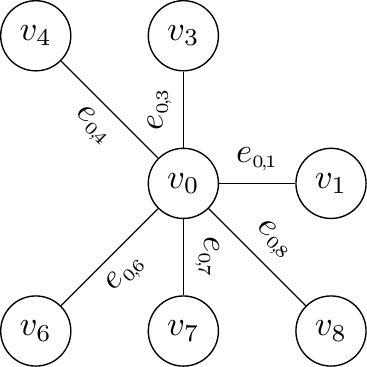}
    \caption{Partially connected graph.}
        \label{fig:partial-graph}
    \end{subfigure}
    \caption{Graph nodes, edges, and costs with 8-connected logic.}
    \label{fig:graph-connections}
\end{figure}

Graph edges can be created for all neighboring nodes as defined by connection logic $\delta_c$. For an \emph{8-connected} logic, any node $v_0$ has potential neighbors $v_1, v_2, \ldots, v_8$ as shown in Fig. \ref{fig:full-graph}, with non-diagonal (odd) and diagonal (even) edges. Due to obstacles, not all neighbors might be reachable, as shown in Fig. \ref{fig:partial-graph} where we assume $v_2, v_5 \in \mathcal{C}_{obs}$ for demonstration purposes.

Accounting for obstacles, all feasible graph edges can be computed as follows:
\begin{equation}
    E = \left \{ e_{m,n} = (\mathcal{Q}_m, \mathcal{Q}_n) \in {V \choose 2} \right \}
\end{equation}
where $i$ and $j$ serve as node identifiers or IDs.

Given graph $\mathcal{G} = (V,E)$, the start $\mathcal{Q}_S$ and goal $\mathcal{Q}_G$ nodes are matched to the closest nodes in $\mathcal{G}$ with labels assigned accordingly. An optimal path is then constructed using \astar{} search on $\mathcal{G}$.
To optimize path construction, \astar{} uses the total cost $f(\mathcal{Q}_n) = g(\mathcal{Q}_n) + h(\mathcal{Q}_n)$ where $g(\mathcal{Q}_n)$ is the cumulative \emph{cost-so-far} from $\mathcal{Q}_{S}$ to $\mathcal{Q}_n$, and $h(\mathcal{Q}_n)$ estimates \emph{cost-to-go}. Similar to Eq. \ref{eq:straight-path-cost}, $g(\mathcal{Q}_n)$ is given by:
\begin{equation}
    g(\mathcal{Q}_{n}) =  g(\mathcal{Q}_m) + c(\mathcal{Q}_{m}, \mathcal{Q}_{n})
    \label{eq:g_cost_fun}
\end{equation}
where $\mathcal{Q}_m$ is the parent node of $\mathcal{Q}_n$, and Eq. \ref{eq:c_cost_fun} calculates function $c(\cdot)$.

Built on the underlying optimalty of Dijkstra's algorithm \cite{dijkstra_note_1959}, the \astar{} heuristic function $h(\cdot)$ maintains optimality and improves search efficiency so long as:

\begin{itemize}
    \item $h(\cdot)$ is admissible, i.e., it never overestimates the true \emph{cost-to-go}.
    \item $h(\cdot)$ is consistent, i.e., for any successor configuration $n$, $h(m) \leq c(m, n) + h(n)$, where $c(\cdot)$ is the true cost to travel from $m$ to $n$.
\end{itemize}

Under these conditions, we propose the following novel heuristic  applicable to motion planning with multiple metric maps:
\begin{equation}
    h_{plus}(\mathcal{Q}_i) = w_{0} \hat{d}(\mathcal{Q}_i, \mathcal{Q}_G) + \sum_{j=1}^{k} w_{j} \hat{s}_j(\mathcal{Q}_i)
    \label{eq:h_cost_fun}
\end{equation}
where $\hat{d}(\cdot)$ approximates the remaining distance to the goal and $\hat{s}_j(\mathcal{Q}_i)$ conservatively estimates the cumulative map-based costs for the final path.

The distance function $\hat{d}(\cdot)$ is chosen to be admissible. For an 8-connected uniform grid, octile distance gives the minimum distance between any node pair. Octile distance $d_{oct}$ extends Manhattan distance by allowing for diagonal transitions. The octile distance between two nodes $m,n$ can be computed as:


\begin{equation}
    \hat{d}(m,n) = d_{oct}(m,n) = \delta_r (\lvert dx - dy \rvert + \sqrt{2} \min(dx, dy))
\end{equation}
where $dx = \lvert n_x - m_x \rvert$, $dy = \lvert n_y - m_y \rvert$ represent the number of horizontal $dx$ and vertical $dy$ steps through the map of resolution $\delta_r$ required to reach node $n$ from $m$. 

Next, using the information encoded by each map in $\mathcal{H}_{\delta_r, \delta_z}$ we estimate the minimum map-based costs for any path to $\mathcal{Q}_G$. From a current node $\mathcal{Q}_i$ an axis-aligned bounding box (AABB) $\mathcal{B}$ is constructed such that:

\begin{equation}
    \mathcal{B} = \begin{pmatrix}
        \min(\mathcal{Q}_{i,x}, \mathcal{Q}_{G,x}) \\
        \min(\mathcal{Q}_{i,y}, \mathcal{Q}_{G,y}) \\
        \max(\mathcal{Q}_{i,x}, \mathcal{Q}_{G,x}) \\
        \max(\mathcal{Q}_{i,y}, \mathcal{Q}_{G,y}) 
    \end{pmatrix}
    = \begin{pmatrix}
        \mathcal{B}_{x,min} \\
        \mathcal{B}_{y,min} \\
        \mathcal{B}_{x,max} \\
        \mathcal{B}_{y,max}
        \end{pmatrix}
\end{equation}
with $ n_{r} = \text{abs}(\mathcal{B}_{y,max} - \mathcal{B}_{y,min}) / \delta_r$ rows and $ n_{c} = \text{abs}(\mathcal{B}_{x,max} - \mathcal{B}_{x,min}) / \delta_r$ columns.

The column and row index mappings between $\mathcal{B}$ and $\mathcal{L}$ are computed as follows:
\begin{equation}
     \mathcal{B}_{l,idx} = \left \lfloor \frac{\mathcal{B}_{y,min} + i \delta_r - \mathcal{L}_{y,min}}{\delta_r} \right \rfloor \qquad \mathcal{B}_{l,idy} = \left \lfloor \frac{\mathcal{B}_{x,min} + j \delta_r - \mathcal{L}_{x,min}}{\delta_r} \right \rfloor
\end{equation}
for $i = 0, 1, \ldots, n_c$ and $j = 0,1, \ldots n_r$.

Using the index bounds for rows $(\mathcal{B}_{0,idy}, \mathcal{B}_{n_r,idy})$ and columns $(\mathcal{B}_{0,idx}, \mathcal{B}_{n_c,idx})$ the $i$th row or $j$th column used by the heuristic can be expressed as:
\begin{align}
    \mathcal{C}_{k,j} &= \{ \mathcal{H}_k(\mathcal{B}_{0,idy}, \mathcal{B}_{j,idx}), \mathcal{H}_k(\mathcal{B}_{1,idy}, \mathcal{B}_{j,idx}), \ldots, \mathcal{H}_k(\mathcal{B}_{n_r,idy}, \mathcal{B}_{j,idx})\} \\
    \mathcal{R}_{k,i} &= \{ \mathcal{H}_k(\mathcal{B}_{i,idy}, \mathcal{B}_{0,idx}), \mathcal{H}_k(\mathcal{B}_{i,idy}, \mathcal{B}_{1,idx}), \ldots, \mathcal{H}_k(\mathcal{B}_{i,idy}, \mathcal{B}_{n_c,idx})\}
\end{align}
for the $k$th cost map in $\mathcal{H}_{\delta_r, \delta_z}$, i.e., for $k \geq 1$.

The minimum cost for the $k$th map-based metric is computed as follows:
\begin{equation}
    \hat{s}_k(n) = \max \left (\sum_{i=1}^{dx} \min(\mathcal{C}_{k,i}), \sum_{i=1}^{dy} \min(\mathcal{R}_{k,i} ) \right )
\end{equation}

By construction, this portion of the heuristic is consistent and admissible. Since both portions of the heuristic are admissible, the overall presented heuristic is admissible as well, guaranteeing \astar{} solution optimality. To test this heuristic, two \astar{} variants are studied in this paper. \astar{dist} uses a traditional Euclidean distance-to-goal heuristic $h_{dist}$ while \astar{plus} applies the novel $h_{plus}$ defined in Eq. \ref{eq:h_cost_fun}.

\subsection{Sampling-based Planning: \bitstar{}}

Batch Informed Trees (\bitstar{}) \cite{gammell_batch_2015} is a sampling-based search algorithm that improves scalability relative to classical graph-based techniques. Extending on previous work \cite{gammell_informed_2014}, \bitstar{} utilizes an iterative search graph $\mathcal{G}$ informed by previous solutions. When a solution is found, \bitstar{} reduces its search space $\mathcal{C}_{free}$, prunes and reuses its search graph, generates a new set of samples in the new $\mathcal{C}_{free}$, and restarts its search. \bitstar{} terminates when a cost threshold has been met or all batches are complete.

For this investigation, the \bitstar{} motion planning problem is defined by:
\begin{itemize}
    \item \emph{Parameters}: $\boldsymbol{\Lambda}_{_{\text{BIT}^*}} = (\mathcal{L}, \mathcal{Q}_S, \mathcal{Q}_G, \mathcal{H}_{\delta_r, \delta_z}, \mathcal{W}, \delta_z, \delta_r, \delta_b, \delta_s)$
    \item \emph{Search Tree}: $\mathcal{T}_{i}$ = \bitstar{} $(\mathcal{T}_{i-1}, \mathcal{H}_{obs}, \delta_s)$ for $i = 1, 2, \ldots, \delta_b$
    \item \emph{Total Cost Function}: $f(\cdot)= g(\cdot) + h(\cdot)$
\end{itemize}
where \bitstar{}$(\cdot)$ returns a graph, and path if found, updated with $\delta_s$ samples per batch, for $\delta_b$ batches/iterations.

\begin{figure}[h]
    \centering
    \begin{subfigure}[t]{.24\textwidth}
		\centering
    \includegraphics[width=\textwidth]{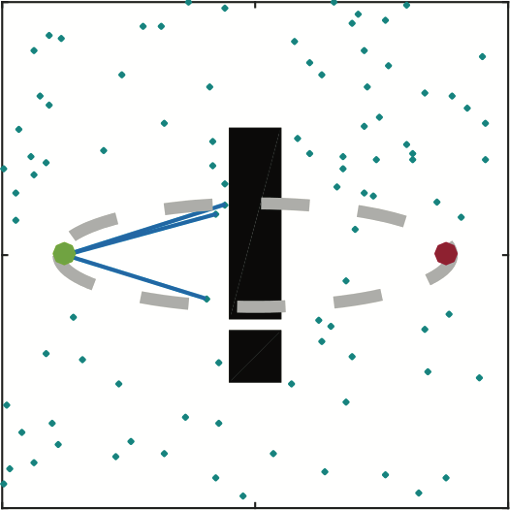}
    \caption{For each batch, the search expands out from the minimum solution.}
    \end{subfigure}
    \hfill
    \begin{subfigure}[t]{.24\textwidth}
		\centering
    \includegraphics[width=\textwidth]{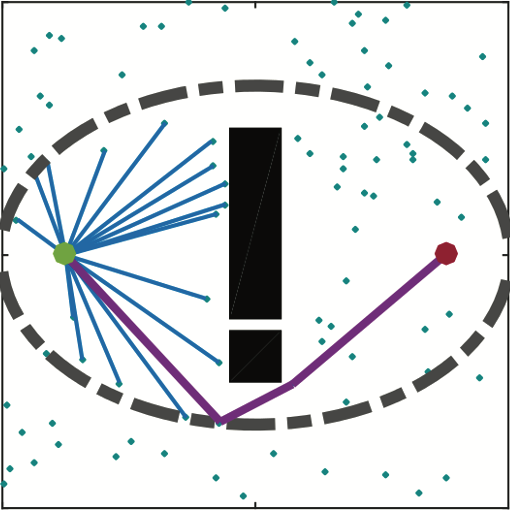}
    \caption{When a solution is found, the batch finishes and a new search space is defined.}
    \end{subfigure}
    \hfill
    \begin{subfigure}[t]{.24\textwidth}
		\centering
    \includegraphics[width=\textwidth]{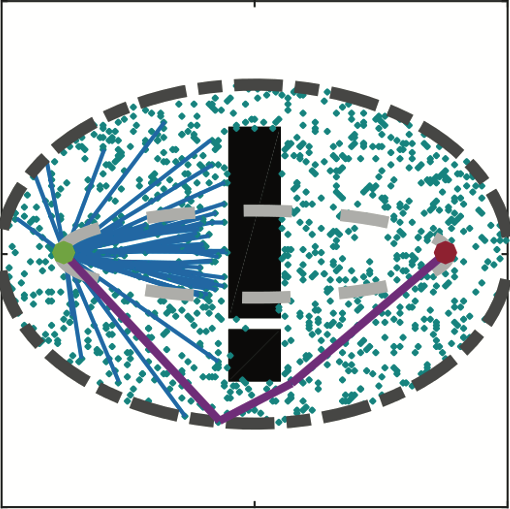}
    \caption{A new batch of samples is added to a newly reduced search space and restarts.}
    \end{subfigure}
    \hfill
    \begin{subfigure}[t]{.24\textwidth}
		\centering
    \includegraphics[width=\textwidth]{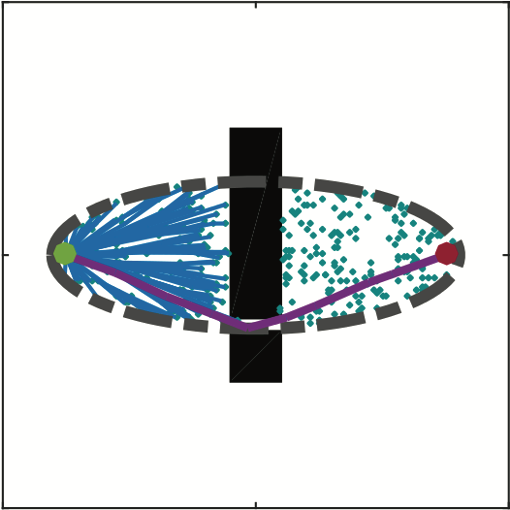}
    \caption{The process repeats to find a better solution every batch.}
    \end{subfigure}
    \caption{\bitstar{} batch process as adapted from \cite{gammell_batch_2015}.}
\label{fig:bit_star_progress}
\end{figure}

Similar to \astar{}, \bitstar{} uses a \emph{cost-so-far} function $g(\cdot)$ and \emph{cost-to-go} heuristic $h(\cdot)$ to search a series of increasingly dense implicit rapidly-exploring random graphs (RRGs) efficiently as illustrated in Fig. \ref{fig:bit_star_progress}, adapted from \cite{gammell_batch_2015}. When initializing the $i$th batch, the search for a solution expands outward from the minimum cost solution, adding feasible connections from $\mathcal{C}_{free,i}$ to a growing tree $\mathcal{T}_i$ with nodes and edges $(V_i, E_i)$. If a solution is found, the batch ends and the search space $\mathcal{C}_{free, i+1}$ is redefined so new samples can only improve the current solution. The previous tree is pruned of any nodes and edges outside of $\mathcal{C}_{free, i+1}$ such that:
\begin{equation}
V_{i+1} = V_i \cap \mathcal{C}_{free,i+1} \qquad E_{i+1} = \left \{ e_{m,n} \; | \; e_{m,n} \in E_{i} \wedge \mathcal{Q}_{m}, \mathcal{Q}_{n} \in V_{i+1} \right \}
\end{equation}
A new set of $\delta_s$ nodes is sampled in $\mathcal{C}_{free, i+1}$, and the search restarts for the next batch. \bitstar{} terminates when all $\delta_b$ batches are complete or the latest solution meets some cost-ending criteria, e.g., a percent change or total cost threshold.

During the first batch, $\mathcal{T}_1$ is initiated such that $V = \{ \mathcal{Q}_S \}$ and $E = \emptyset$. Nodes are added to the closest node in the current tree if a collision-free edge is feasible and they improve the best solution so far $\hat{\zeta}$. The costs of of adding a new node $\mathcal{Q}_n$ with an edge $e_{m,n}$ are computed using Eq. \ref{eq:g_cost_fun} for $g(\mathcal{Q}_n)$ and Eq. \ref{eq:c_cost_fun} for $c(\mathcal{Q}_m, \mathcal{Q}_n$). Similar to the \astar{} variants, \bitstar{dist} uses $h_{dist}$ as its heuristic while \bitstar{plus} applies the novel $h_{plus}$ defined in Eq. \ref{eq:h_cost_fun}.

\section{Manhattan Metric Map Results}
\label{sec:results}

\begin{figure}[h]
    \centering
    \begin{subfigure}{0.32\textwidth}
        \centering
        \includegraphics[width=0.99\columnwidth]{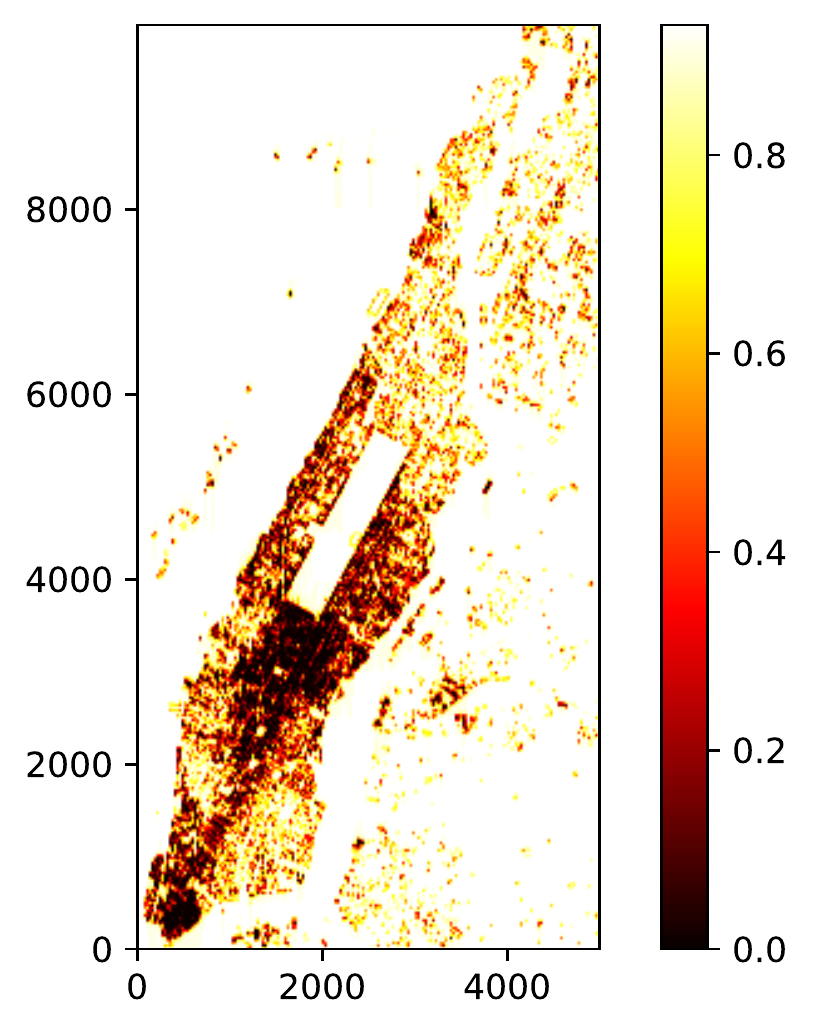}
        \caption{GPS 2m res map at 20m.}
    \end{subfigure}
    \hfill
    \begin{subfigure}{0.32\textwidth}
        \centering
        \includegraphics[width=0.99\columnwidth]{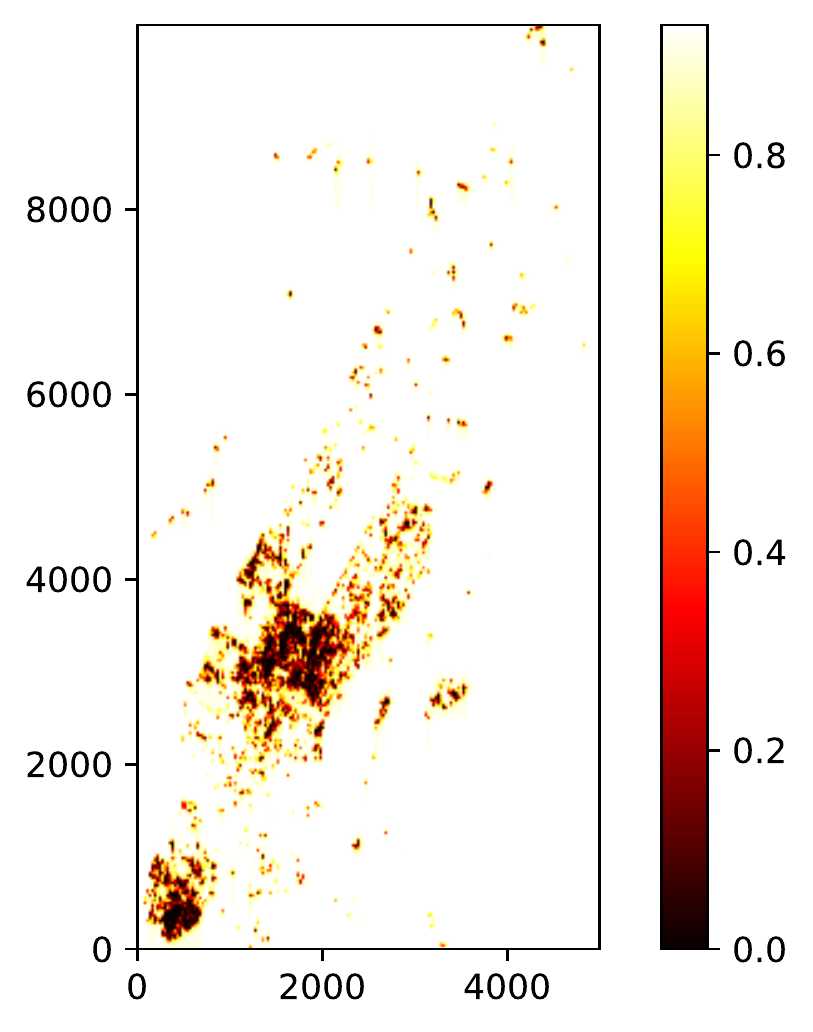}
        \caption{GPS 2m res map at 60m.}
    \end{subfigure}
    \hfill
    \begin{subfigure}{0.32\textwidth}
        \centering
        \includegraphics[width=0.99\columnwidth]{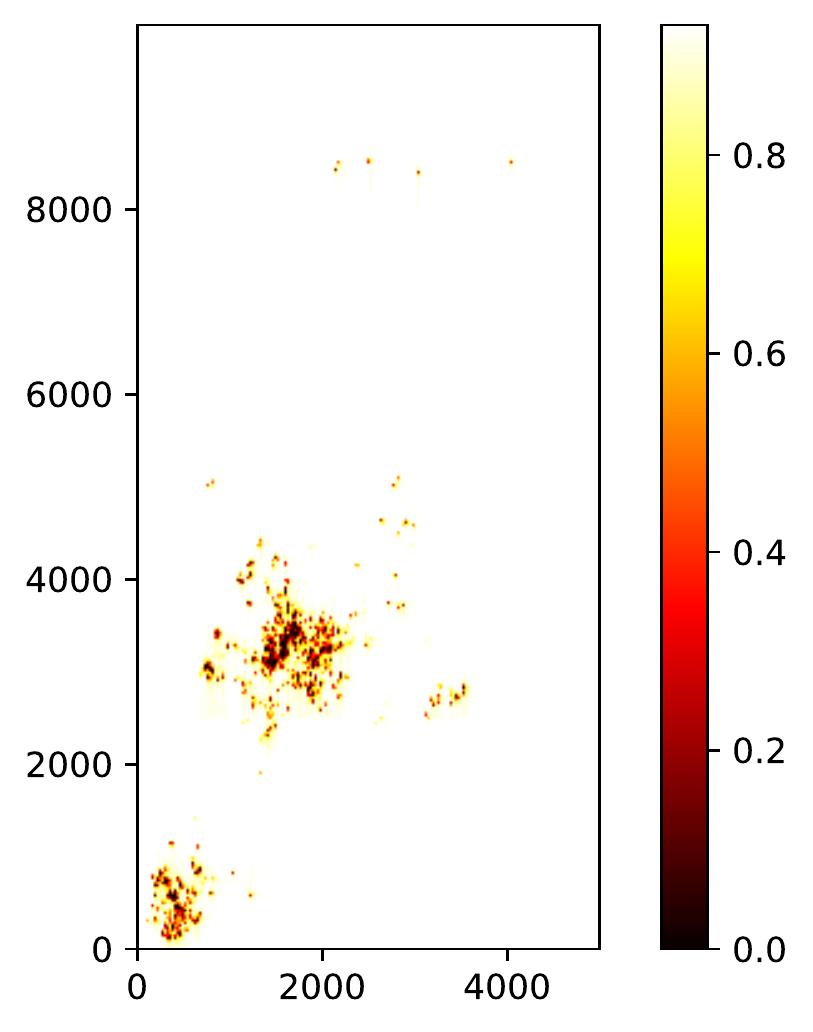}
        \caption{GPS 2m res map at 122m.}
    \end{subfigure}
    \caption{GPS metric maps for low, medium, and high-altitude urban flight.}
    \label{fig:low_med_high_gps}
\end{figure}

Metric maps over Manhattan region $\mathcal{L}$ at three different resolutions (2m, 5m, and 10m) were generated for four small UAS AGL flight altitudes: 20m (low-altitude), 60m (medium-altitude), 122m (high-altitude), and 600m (ceiling-altitude). This altitude set covers sUAS flight paths that range from deep inside the New York City urban canyon (low-altitude) to above all buildings (ceiling-altitude). 
Fig. \ref{fig:low_med_high_gps} shows GPS maps for low (20m), medium (60m), and high (122m) altitude flight. GPS metric scores are normalized between 0 and 1, where $m_{gps}=1$ indicates the highest accuracy. As expected, GPS accuracy is highest in building-free areas, i.e., the Hudson River or Central Park, or residential areas with single-family homes, i.e., New Jersey. GPS accuracy decreases in low-altitude urban canyon regions with tall buildings. 

For medium-altitude flight, the effects of urban canyon flight lessen. Upper Manhattan and Brooklyn (lower right) are now areas with high GPS accuracy. Similarly, high GPS accuracy areas now appear in Lower Manhattan but to a lesser extent. The Financial District (bottom left) and Midtown Manhattan (below Central Park) still include low GPS accuracy regions. This is to be expected as these areas are known for their tall buildings, e.g., One World Trade Center and Central Park Tower. The UAS primarily operates above the urban canyon at high and ceiling flight altitudes with near-perfect GPS accuracy. 

\begin{figure}[h]
    \centering
    \begin{subfigure}{0.32\textwidth}
        \centering
        \includegraphics[width=0.99\columnwidth]{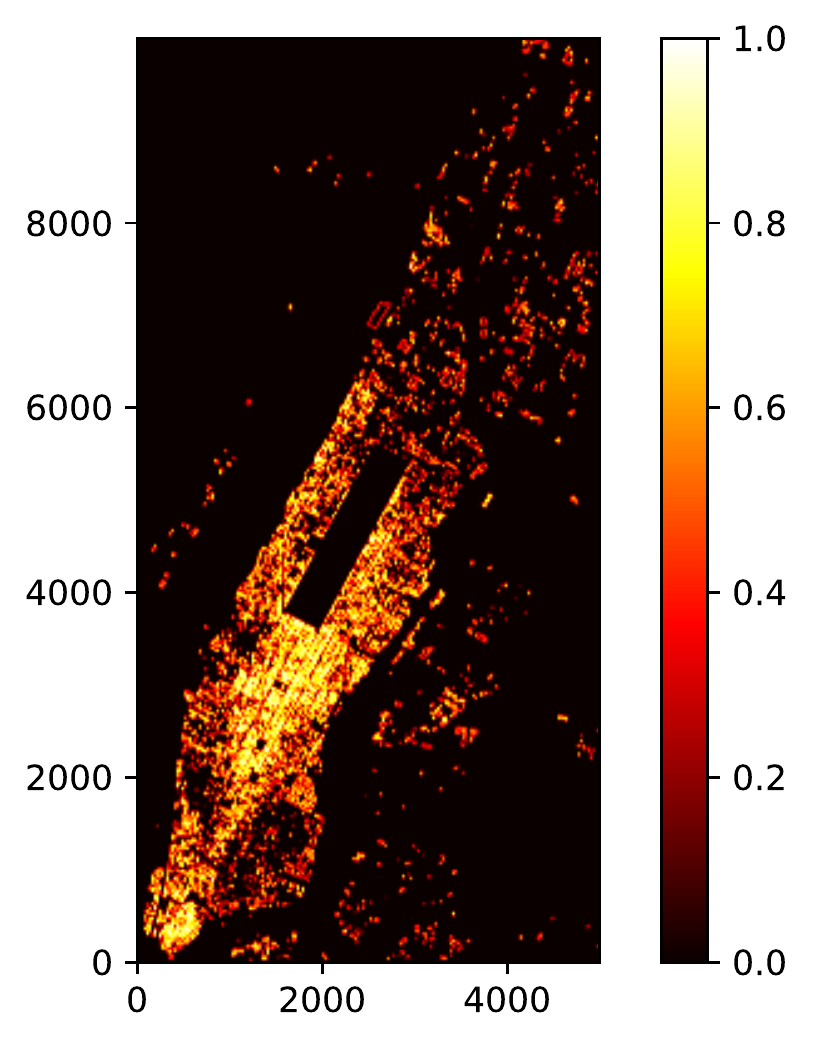}
        \caption{Lidar 2m res map at 20m.}
    \end{subfigure}
    \hfill
    \begin{subfigure}{0.32\textwidth}
        \centering
        \includegraphics[width=0.99\columnwidth]{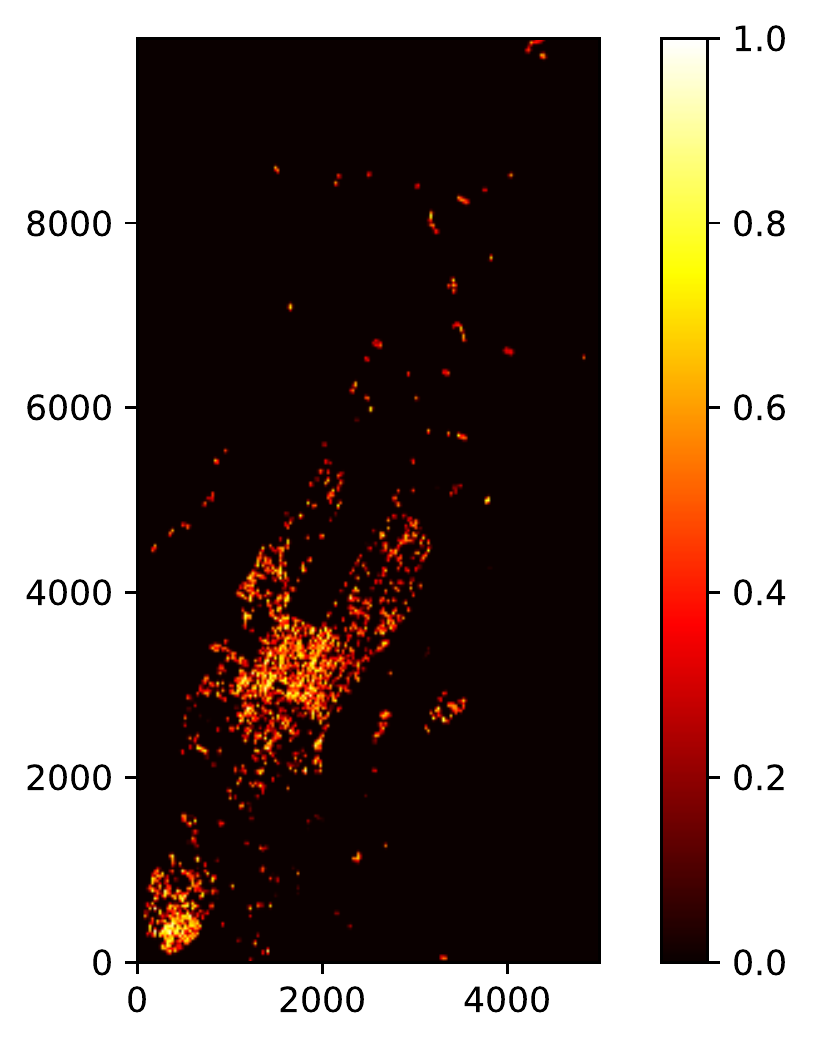}
        \caption{Lidar 2m res map at 60m.}
    \end{subfigure}
    \hfill
    \begin{subfigure}{0.33\textwidth}
        \centering
        \includegraphics[width=0.96\columnwidth]{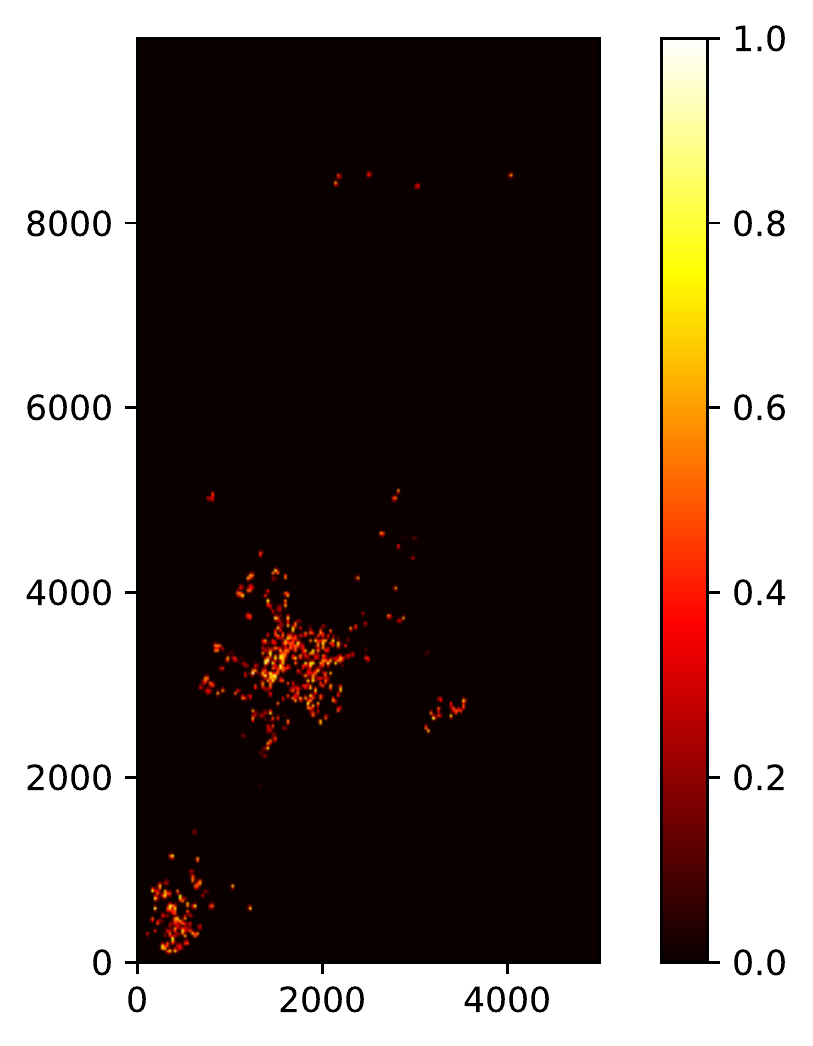}
        \caption{Lidar 2m res map at 122m.}
    \end{subfigure}
    \caption{Lidar metric maps for low, medium, and high-altitude urban flight.}
    \label{fig:low_med_lidar}
\end{figure}

Fig. \ref{fig:low_med_lidar} shows expected lidar performance for low-altitude and medium-altitude flight. In contrast to GPS, lidar performance is better at lower altitudes since the urban canyon offers in-range point cloud data and better visibility of its surroundings. In low-altitude flight, lidar performance is highest in the East Side, West Side, Midtown, and Downtown Manhattan areas densely packed with commercial and tourist high-rises. Weak lidar returns can be found in Uptown Manhattan, New Jersey, Brooklyn, and Queens, areas with mostly low-rise and residential buildings.

Medium-altitude lidar analysis shows a significant drop in performance. Of the four predominant high $m_{lidar}$ regions from the low-altitude analysis, only Midtown Manhattan remains. A pattern emerges at this altitude that suggests the potential for GPS to complement lidar, and vice-versa. Areas of low $m_{gps}$ due to the urban canyon coexist with high $m_{lidar}$ areas, and low $m_{lidar}$ due to the absence of nearby obstacles results in high $m_{gps}$ areas without satellite obstruction. This effect becomes more apparent at high-altitude flight and above.

\begin{figure}[h]
    \centering
    \begin{subfigure}{0.49\textwidth}
        \centering
        \includegraphics[width=0.65\columnwidth]{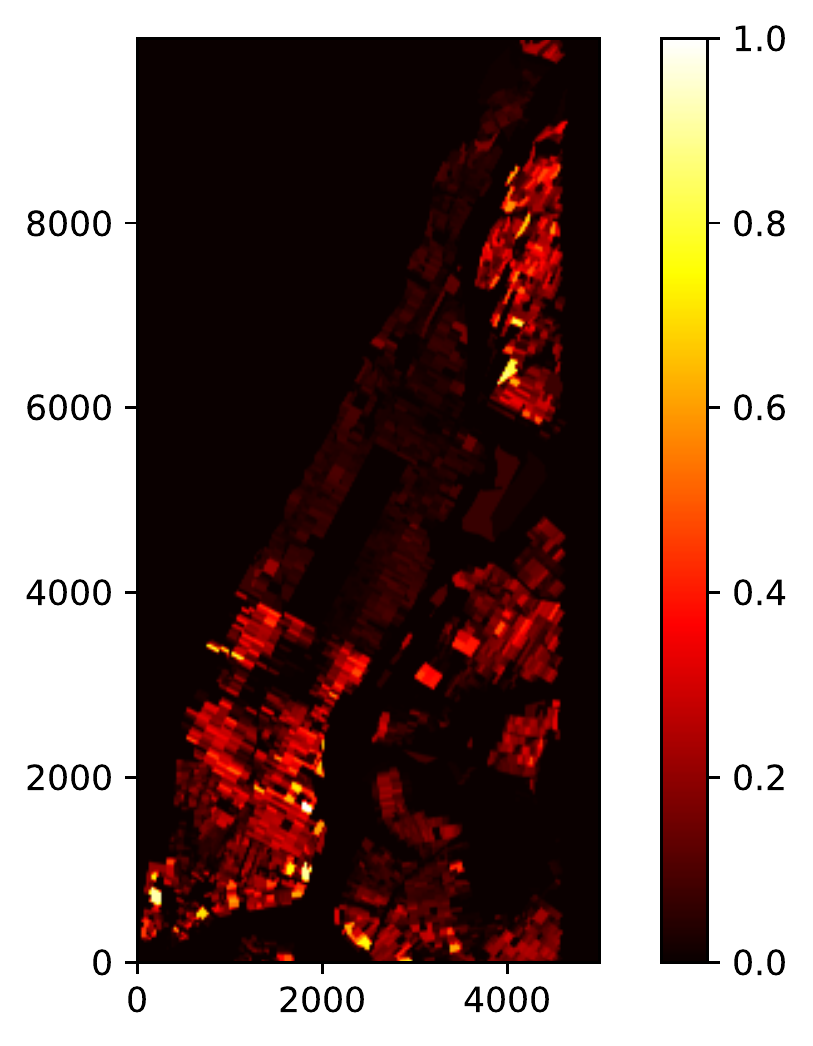}
        \caption{Daytime population (2m res)}
    \end{subfigure}
    \hfill
    \begin{subfigure}{0.49\textwidth}
        \centering
        \includegraphics[width=0.65\columnwidth]{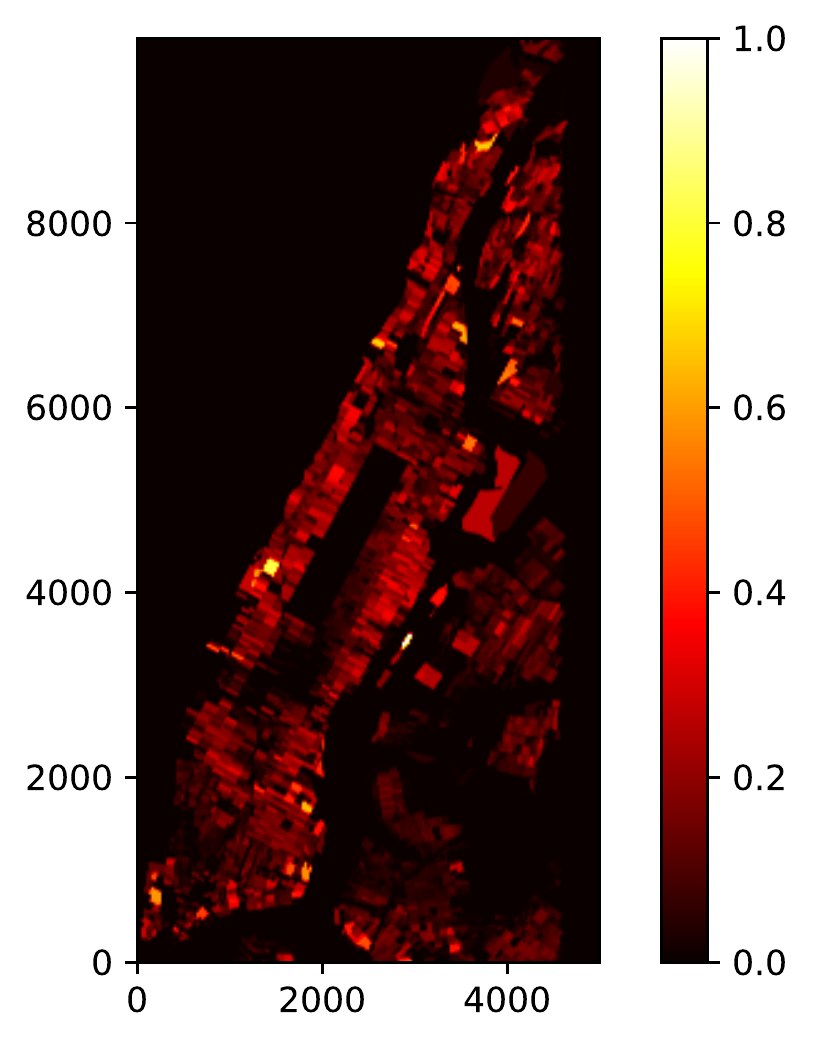}
        \caption{Population at night (2m res)}
    \end{subfigure}
    \caption{Population metric maps over Manhattan for day and night hours.}
    \label{fig:population_maps}
\end{figure}

Day and night population metric maps, shown in Fig. \ref{fig:population_maps}, are independent of flight altitude. The following daytime population scaling factors were used:  $\Gamma_{comm} = 3.0$ and $\Gamma_{resi} = 0.5$.  These values are biased toward a net population influx into Manhattan for the workday as show in Table \ref{tab:pop_modified}. The population map results validate the expected residence-to-work and work-to-residence commuting patterns and constraints discussed in Sec. \ref{sec:mm-pop-maps}.

\begin{table}[h]
    \caption{Work weekday and nighttime population estimates in millions.}
    \label{tab:pop_modified}
    \centering
    \begin{tabular}{c c  c}
    \toprule
                & Residential & Commericial \\
    \midrule
        Daytime & 0.48 & 3.96 \\
        Nighttime & 0.97 & 1.32 \\
    \bottomrule
    \end{tabular}
\end{table}

\begin{figure}[h]
    \centering
    \begin{subfigure}{0.32\textwidth}
        \centering
        \includegraphics[width=0.99\columnwidth]{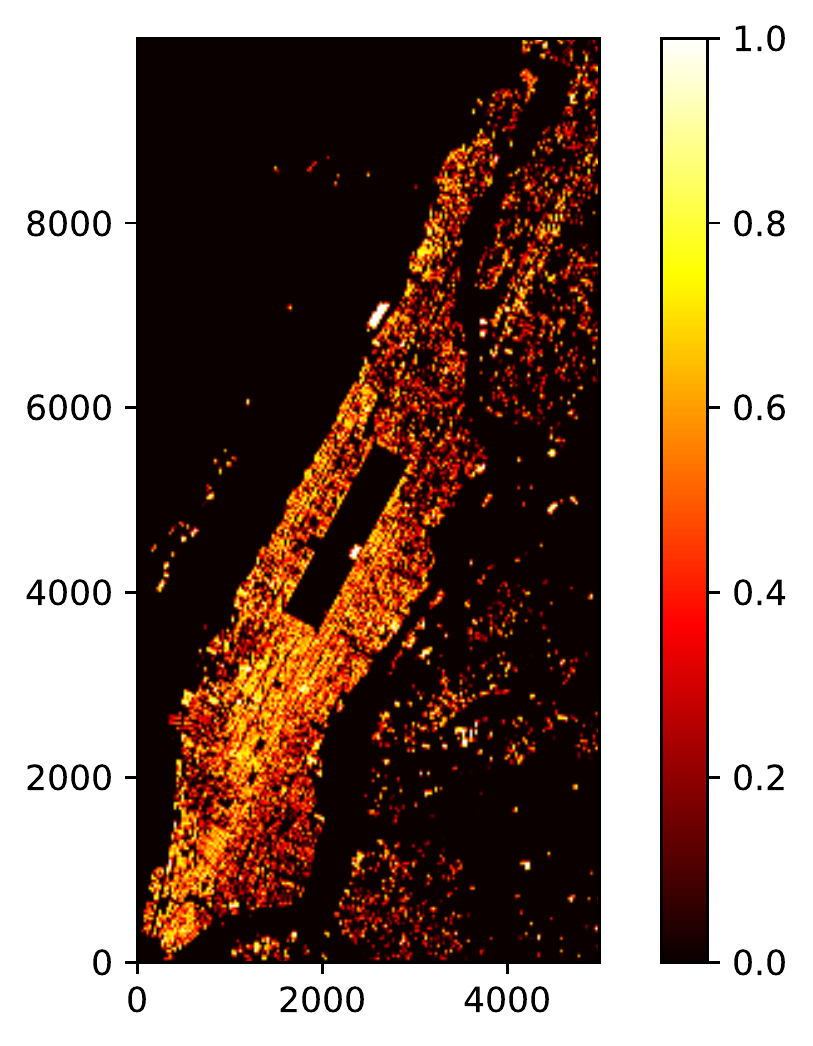}
        \caption{Risk at 2m res at 20m.}
    \end{subfigure}
    \hfill
    \begin{subfigure}{0.32\textwidth}
        \centering
        \includegraphics[width=0.99\columnwidth]{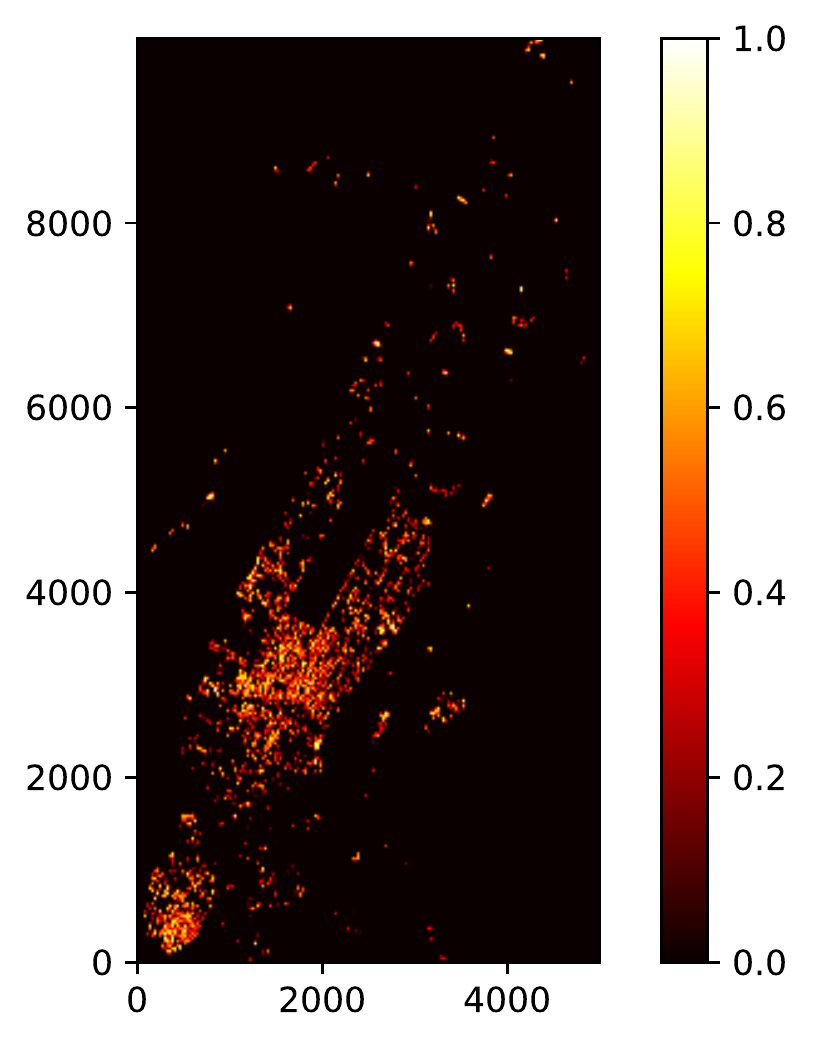}
        \caption{Risk at 2m res at 60m.}
    \end{subfigure}
    \hfill
    \begin{subfigure}{0.33\textwidth}
        \centering
        \includegraphics[width=0.96\columnwidth]{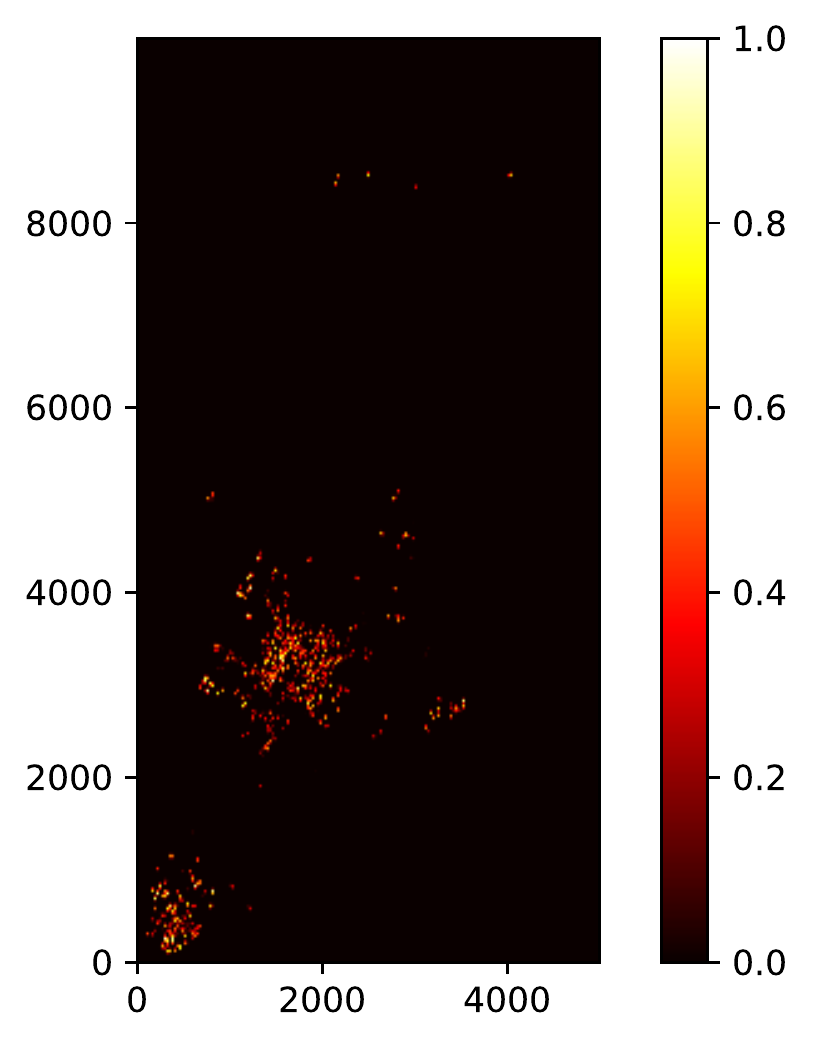}
        \caption{Risk at 2m res at 122m.}
    \end{subfigure}
    \caption{Proximity risk metric maps for low-altitude and medium-altitude flight.}
    \label{fig:low_med_high_risk}
\end{figure}

Proximity risk maps identify obstacle-free map grid points with decaying risk value over a distance $d_{thresh}$ around buildings, the risk is one at the building, linearly decreasing to 0 at $d_{thresh}$. High proximity risk areas are mostly in the Manhattan borough, as shown in Fig. \ref{fig:low_med_high_risk}. For low altitude-flight, except for the Hudson River, New Jersey, and Central Park, a building can be found within 10m in most grids. Large portions of the Bronx, Queens, Brooklyn, and Uptown Manhattan become risk-free zones at medium-altitude flight. Only Downtown and Midtown Manhattan remain at high-altitude flight due to the congestion of tall buildings, as discussed earlier.

\section{Monte Carlo Simulation Procedure}
\label{sec:sim}

All map generation and planning simulations were performed using the Google Cloud: Compute Engine (CE). Maps and Monte Carlo planning simulations were generated using ten \emph{n1-standard-16} virtual machines (VMs). Two geospatial datasets were used for all simulations: (1) OSM and (2) TIGER. OSM data was downloaded from PlanetOSM\footnote[2]{https://planet.openstreetmap.org/} as a 50+ GB PBF file. TIGER\footnote[3]{https://www.census.gov/geographies/mapping-files.html} 2010 US Census data was downloaded directly from the US Census Bureau as a 180+ MB shapefile. The Geospatial Data Abstraction Library (GDAL) was used to uncompress and extract all Manhattan-specific data within $\mathcal{L}$. 
Start $\mathcal{Q}_S$ and goal $\mathcal{Q}_G$ configurations were sampled across $\mathcal{L}$ to capture all relevant subdomains, e.g., flying over water, suburban, and high rise building areas. 
Weighting vectors $\mathcal{W}$ were randomly generated for all problem instances. 
Each motion planning algorithm was implemented as discussed in Sec. \ref{sec:planning-algorithms} in Cython, Python's optimized statically compiled variant. Cython takes advantage of Python's high-level, easily readable syntax while providing speeds comparable to C/C++ on execution. All planning instances were equally distributed among all VMs and ran against each planner.

\section{Path Planning Results}
\label{sec:path-results}

This section analyzes solution path properties from Monte Carlo simulations. Case studies are selected for each altitude  $z^{*} \in $ \{20m, 60m, 122m, 600m\} AGL. Motion planning solutions generated within the allotted time (three minutes) are shown relative to the total unweighted cost map $\mathcal{H}_{total}$ referenced during planning. Total cost maps  $\mathcal{H}_{total}(z^{*})$ are defined by:
\begin{equation}
      \mathcal{H}_{total}(z^{*}) = \mathcal{H}_{gps}(z^{*}) + \mathcal{H}_{lidar}(z^{*}) + \mathcal{H}_{pop}(z^{*}) + \mathcal{H}_{risk}(z^{*}) 
\end{equation}
and normalized using min-max normalization:
\begin{align}
    \mathcal{H}_{shift}(z^{*}) &= \mathcal{H}_{total}(z^{*}) -  \min(\mathcal{H}_{total}(z^{*})) J \\
      \mathcal{H}_{norm}(z^{*}) &= \frac{1}{\max(\mathcal{H}_{shift}(z^{*}))}\mathcal{H}_{shift}(z^{*})
\end{align}
where $J$ is a matrix of ones with the same dimensions as $\mathcal{H}_{total}$. To compare, we focus on daytime population for \{20m, 60m\} AGL flight and nighttime population for \{122m, 600m\} flight. Motion planners that found a solution are labeled on the top-left corner of each map.

\begin{figure}[!htb]
    \centering
    \begin{subfigure}{0.31\textwidth}
        \centering
        \includegraphics[width=0.91\columnwidth]{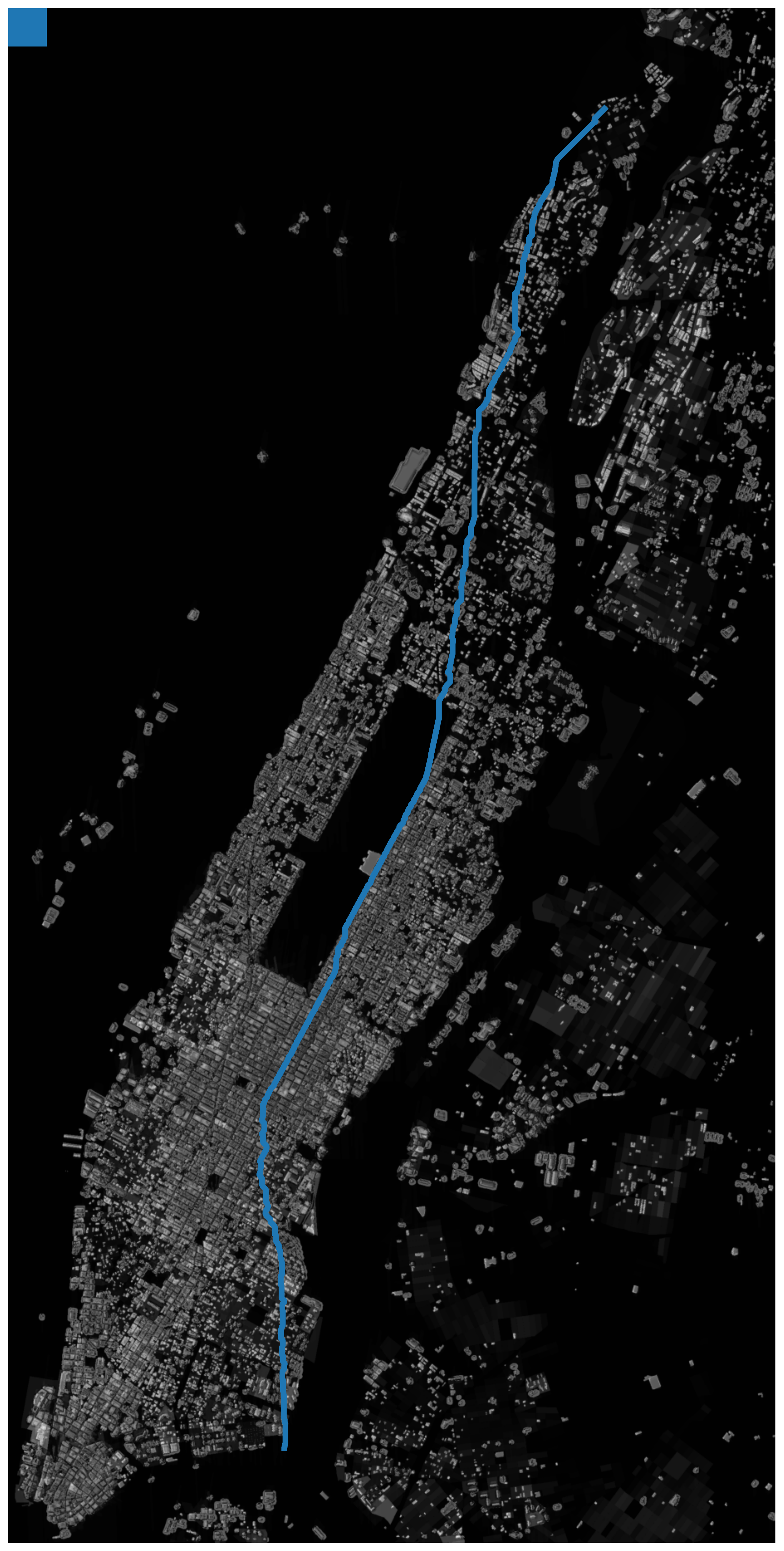}
        \caption{}
        \label{fig:low-alt-case-a}
    \end{subfigure}
    \hfill
    \begin{subfigure}{0.31\textwidth}
        \centering
        \includegraphics[width=0.91\columnwidth]{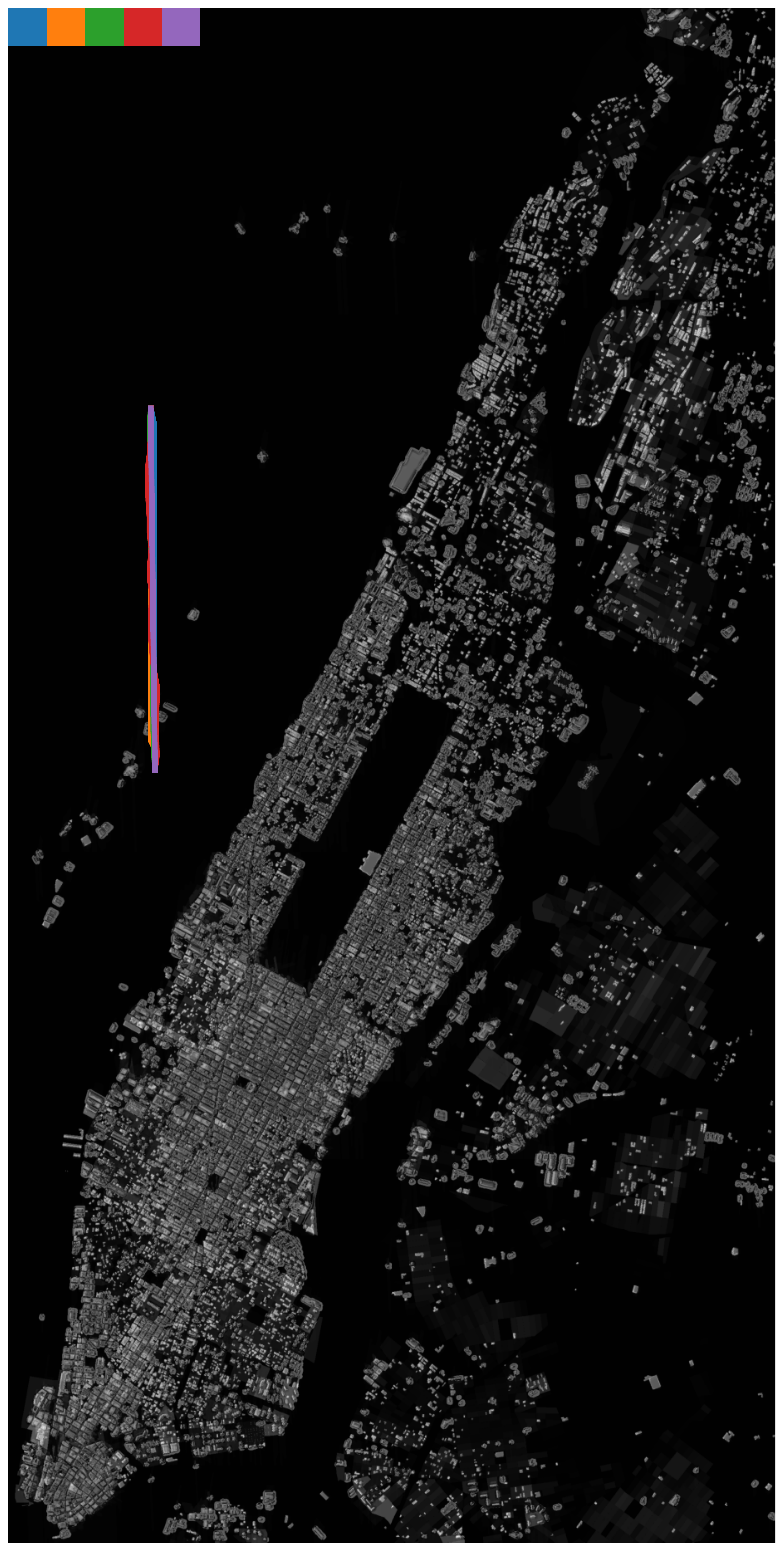}
        \caption{}
        \label{fig:low-alt-case-b}
    \end{subfigure}
    \hfill
    \begin{subfigure}{0.31\textwidth}
        \centering
        \includegraphics[width=0.91\columnwidth]{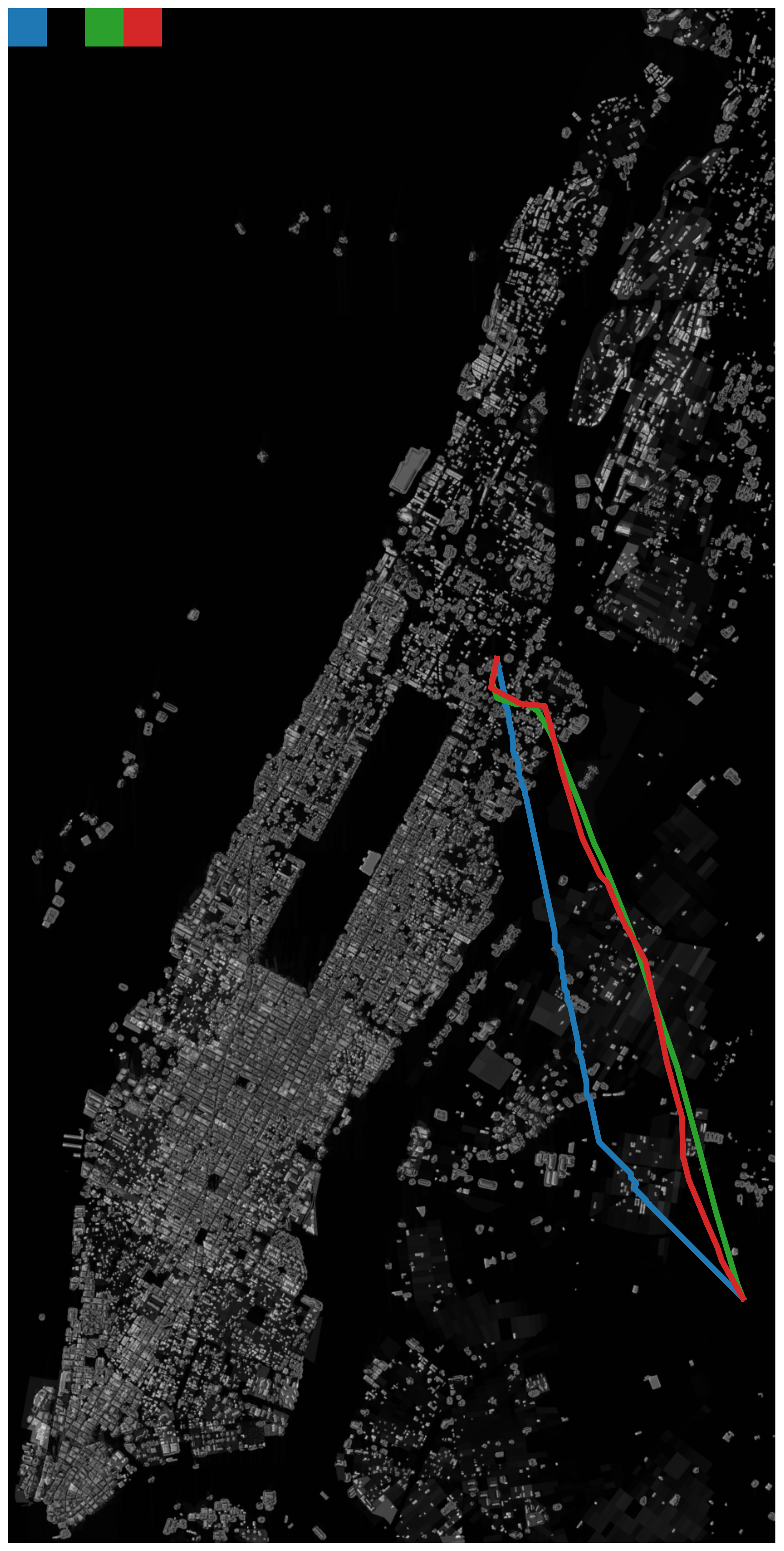}
        \caption{}
        \label{fig:low-alt-case-c}
    \end{subfigure}
    \begin{tabular}{c c c c c}
        \vspace{-09pt}\\
        \astar{dist} \textcolor{tab10-blue}{$\blacksquare$} & \astar{plus} \textcolor{tab10-orange}{$\blacksquare$} & \bitstar{dist} \textcolor{tab10-green}{$\blacksquare$} & \bitstar{plus} \textcolor{tab10-red}{$\blacksquare$} & \ptp{} \textcolor{tab10-purple}{$\blacksquare$} \\
        \vspace{-19pt}\\
    \end{tabular}
    \caption{Example solution paths at 20m AGL, 5m resolution maps in New York City.}
    \label{fig:low_alt_paths}
\end{figure}

For low-altitude flight (20m AGL) obstacle-related costs are prominent in $\mathcal{H}_{norm}$, where $\mathcal{H}_{norm} = 0$ is depicted in black with a gradient to white for $\mathcal{H}_{norm} = 1$ in Fig. \ref{fig:low_alt_paths}. Manhattan, the Bronx, and portions of Queens/Brooklyn display high cost values attributed to tall buildings and urban canyon effects. At such a low altitude, a motion planner requires efficient obstacle-avoidance to find a feasible solution. As shown in Fig. \ref{fig:low-alt-case-a}, for a long-range flight traversing through Manhattan only \astar{dist} was able to find a solution. In contrast, for short-range flights over New Jersey, all planners were able to generate a feasible flight path as shown in Fig. \ref{fig:low-alt-case-b}.  Fig. \ref{fig:low-alt-case-c} shows a mid-range flight with some obstacles present over parts of Queens and Manhattan. The modest number of obstacles allowed three out of the five motion planners to terminate but with different path traits. As described below, \astar{dist} followed a grid-based path that is minimum distance only with respect to that grid, while the \bitstar{} variants took another option that is more direct because \bitstar{} does not rely on the $5m$ resolution map grid apart from estimates of cost. 

\begin{figure}[!htb]
    \centering
    \begin{subfigure}{0.31\textwidth}
        \centering
        \includegraphics[width=0.91\columnwidth]{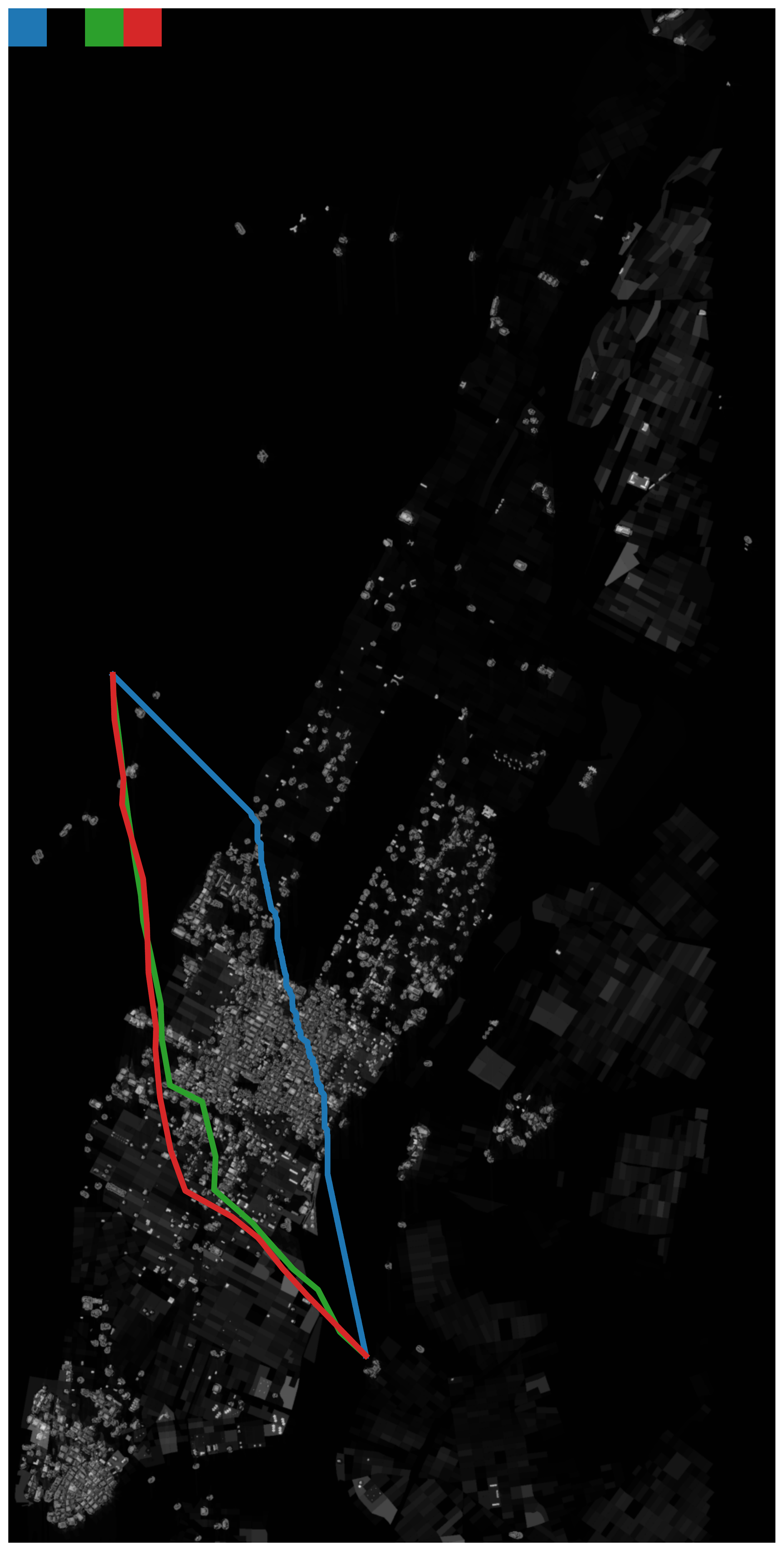}
        \caption{}
        \label{fig:med-alt-case-a}
    \end{subfigure}
    \hfill
    \begin{subfigure}{0.31\textwidth}
        \centering
        \includegraphics[width=0.90\columnwidth]{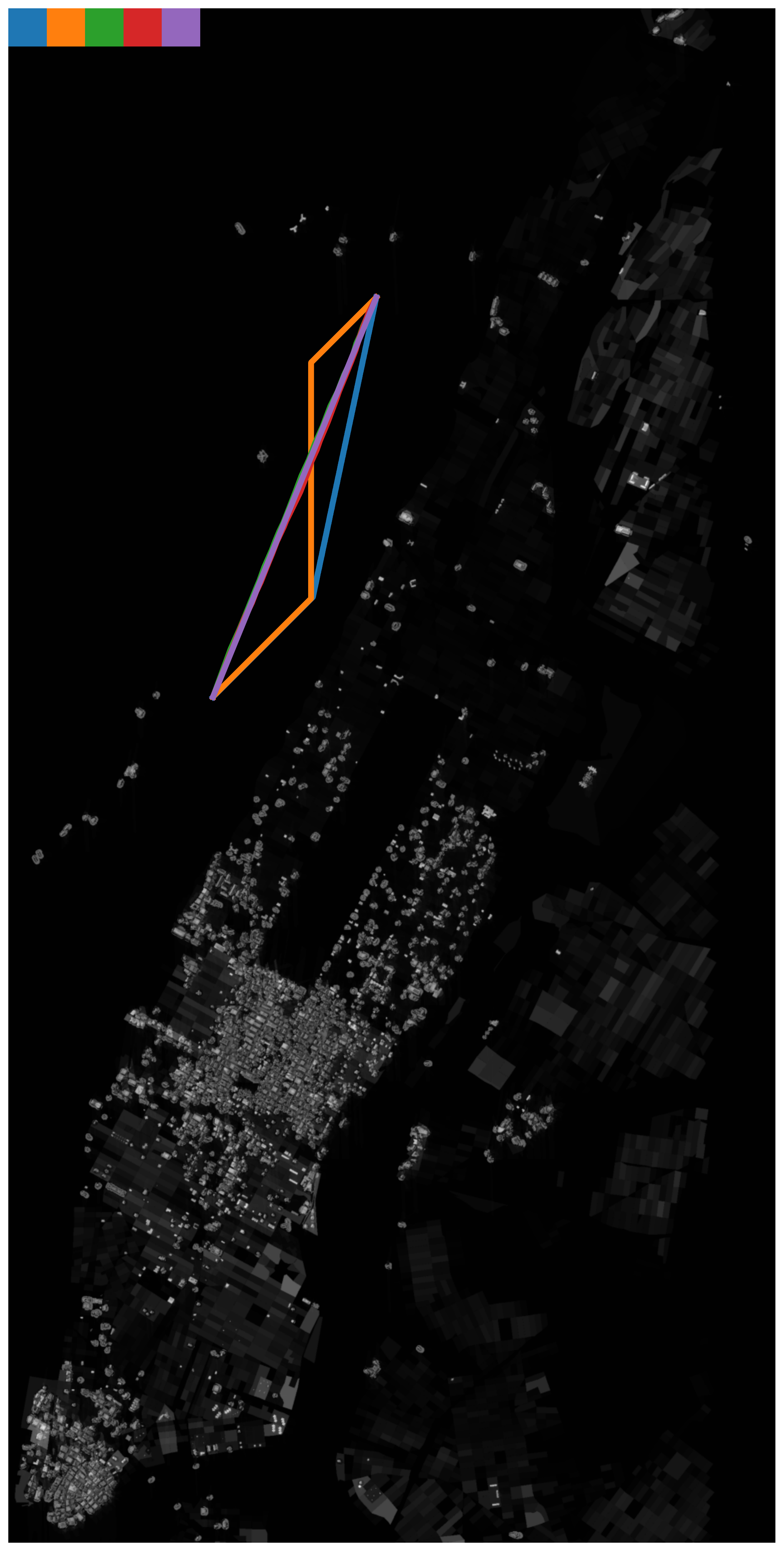}
        \caption{}
        \label{fig:med-alt-case-b}
    \end{subfigure}
    \hfill
    \begin{subfigure}{0.31\textwidth}
        \centering
        \includegraphics[width=0.90\columnwidth]{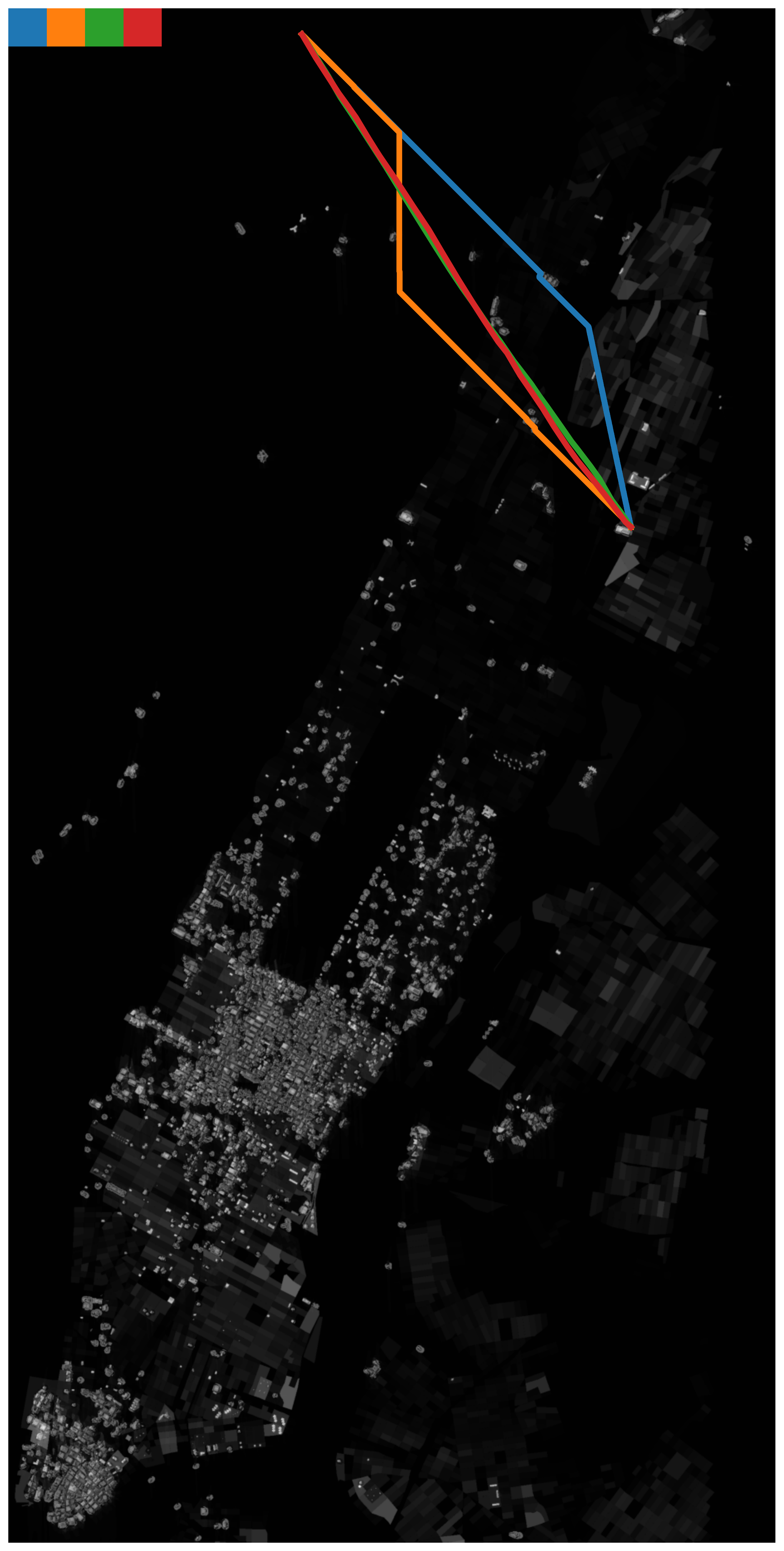}
        \caption{}
        \label{fig:med-alt-case-c}
    \end{subfigure}
    \begin{tabular}{c c c c c}
        \vspace{-09pt}\\
        \astar{dist} \textcolor{tab10-blue}{$\blacksquare$} & \astar{plus} \textcolor{tab10-orange}{$\blacksquare$} & \bitstar{dist} \textcolor{tab10-green}{$\blacksquare$} & \bitstar{plus} \textcolor{tab10-red}{$\blacksquare$} & \ptp{} \textcolor{tab10-purple}{$\blacksquare$} \\
        \vspace{-19pt}\\
    \end{tabular}
    \caption{Example solution paths at 60m AGL, 5m resolution maps in New York City.}
    \label{fig:med_alt_paths}
\end{figure}

For mid-altitude flight (60m AGL), similar path and $\mathcal{H}_{norm}$ characteristics are observed in the Fig. \ref{fig:med_alt_paths} example paths. At this height, obstacles are only present in the Financial District (lower left) and Midtown Manhattan. Population now plays a more significant role in low-rise areas, especially the neighboring boroughs. Fig. \ref{fig:med-alt-case-a} depicts a path attempting to traverse Midtown Manhattan. Motion planners circumvented the dense group of tall buildings with \bitstar{dist} taking ``shortcuts'' to minimize distance while \bitstar{plus} navigates through lower population and risk areas. Fig. \ref{fig:med-alt-case-b} investigates paths generated over the Hudson River. With no population or obstacle-related costs, all motion planners are capable of constructing feasible paths. \bitstar{} variants and \ptp{} take a direct approach from $\mathcal{Q}_{S}$ to $\mathcal{Q}_G$. 
The \astar{} variants follow eight-connected grids.  With the $5m$ resolution case study map, each \astar{} step is either $5m$ along a primary compass direction or $7.07m$ along a $45$ degree diagonal.  This grid-based routing process leads to longer thus higher cost paths compared with direct routes, e.g., a distance cost of $5625m$ for \ptp{} versus $6092m$ for \astar{dist} in the example from Fig. \ref{fig:med-alt-case-b}. This phenomenon is also observed in Fig. \ref{fig:med-alt-case-c}.

\begin{figure}[!htb]
    \centering
    \begin{subfigure}{0.31\textwidth}
        \centering
        \includegraphics[width=0.91\columnwidth]{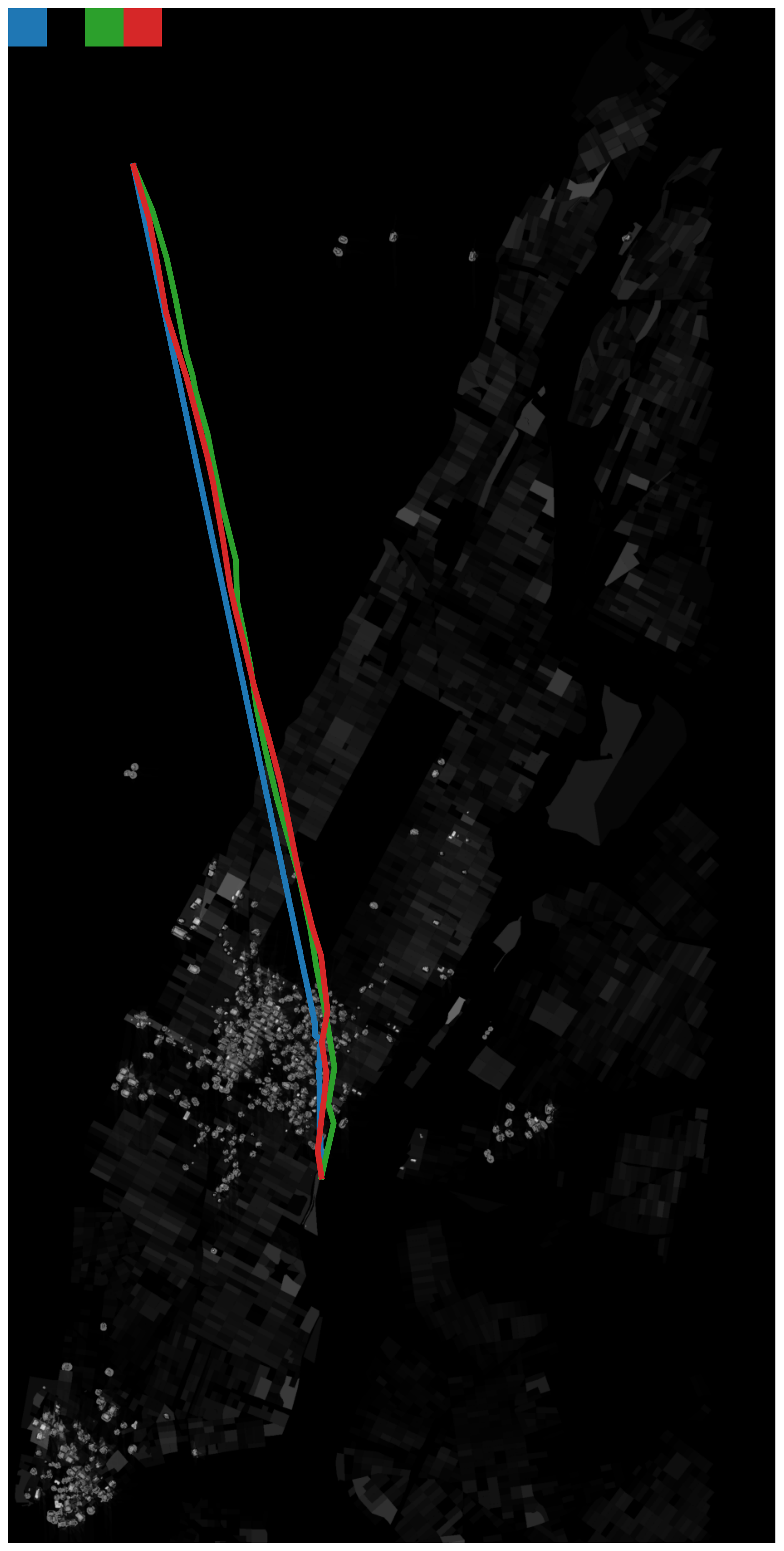}
        \caption{}
         \label{fig:high-alt-case-a}
    \end{subfigure}
    \hfill
    \begin{subfigure}{0.31\textwidth}
        \centering
        \includegraphics[width=0.91\columnwidth]{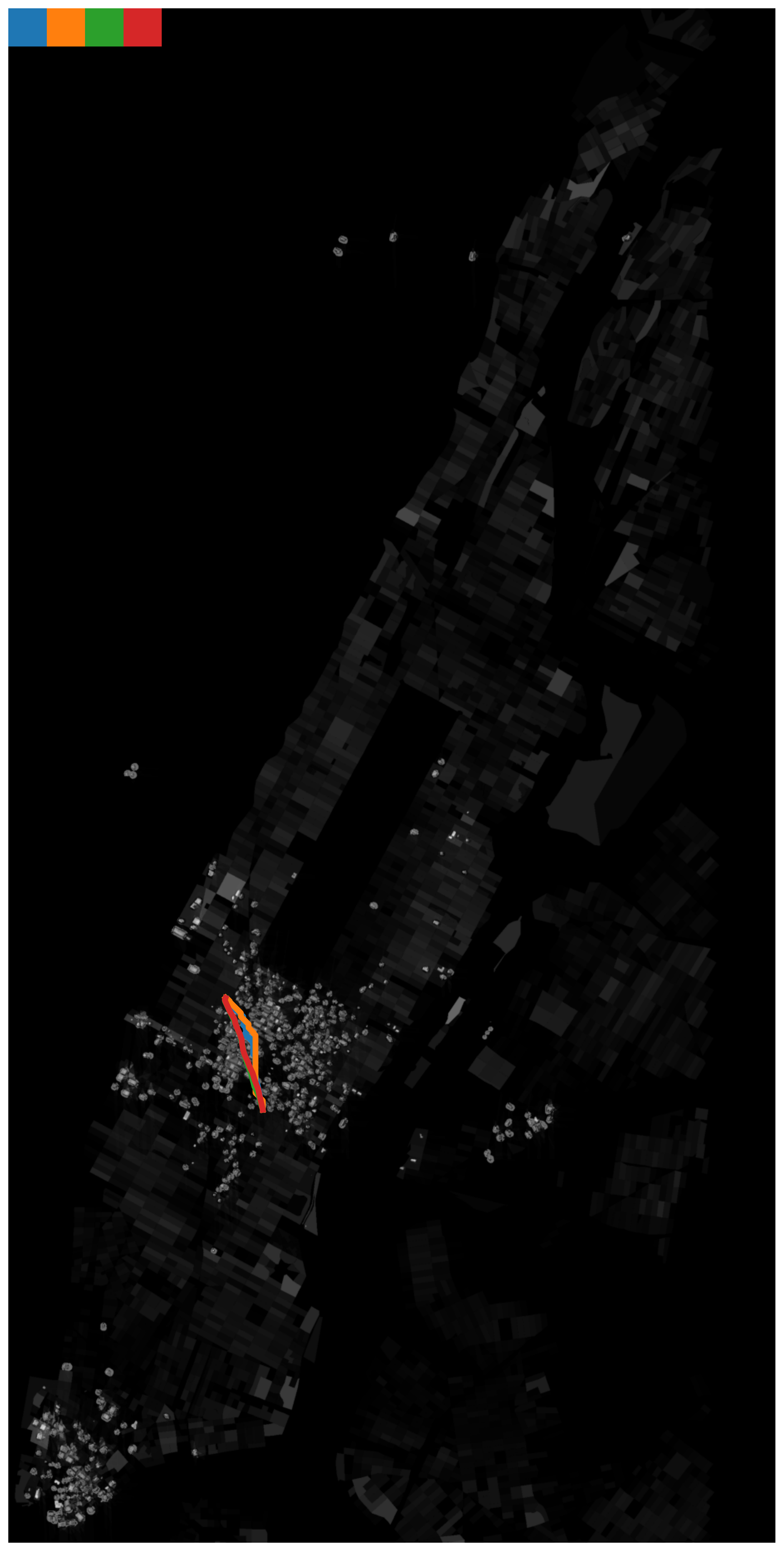}
        \caption{}
        \label{fig:high-alt-case-b}
    \end{subfigure}
    \hfill
    \begin{subfigure}{0.31\textwidth}
        \centering
        \includegraphics[width=0.91\columnwidth]{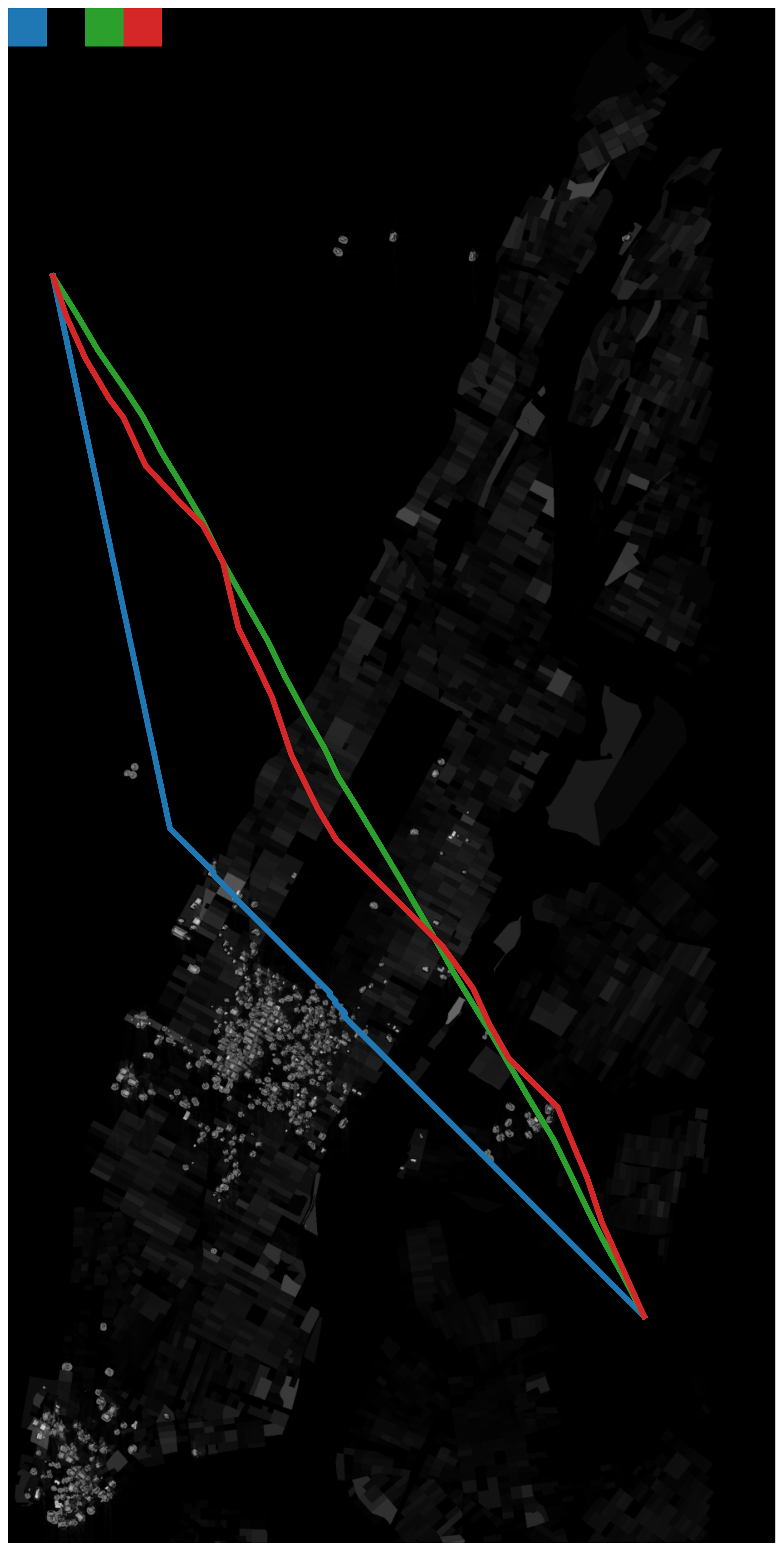}
        \caption{}
        \label{fig:high-alt-case-c}
    \end{subfigure}
    \begin{tabular}{c c c c c}
        \vspace{-09pt}\\
        \astar{dist} \textcolor{tab10-blue}{$\blacksquare$} & \astar{plus} \textcolor{tab10-orange}{$\blacksquare$} & \bitstar{dist} \textcolor{tab10-green}{$\blacksquare$} & \bitstar{plus} \textcolor{tab10-red}{$\blacksquare$} & \ptp{} \textcolor{tab10-purple}{$\blacksquare$} \\
        \vspace{-19pt}\\
    \end{tabular}
    \caption{Example solution paths at 122m AGL, 5m resolution maps in New York City.}
    \label{fig:high_alt_paths}
\end{figure}

For high-altitude flight (122m AGL), tall buildings only remain in highly concentrated areas of the Financial District and Midtown Manhattan. Fig. \ref{fig:high-alt-case-a} and Fig. \ref{fig:high-alt-case-b} illustrate the success of motion planners when flying in these areas for short and long-range flight. In the first case, paths are generated from New Jersey, across the Hudson, and into Midtown Manhattan. Given the long range and abundance of obstacles upon approach, only \astar{dist} and the \bitstar{} variants successfully terminated. However, with a reduced distance between $\mathcal{Q}_S$ and $\mathcal{Q}_G$, \astar{plus} now terminates and takes a safer path than the rest. Furthermore, range can also be an issue for \bitstar{plus}. As shown in Fig. \ref{fig:high-alt-case-c}, \bitstar{dist} and \bitstar{plus} generate noticeably different paths. Given \bitstar{plus} had to search more nodes to minimize non-distance costs, it had fewer batches, or iterations, to return its best-cost solution by the planning deadline.

\begin{figure}[!htb]
    \centering
    \begin{subfigure}{0.31\textwidth}
        \centering
        \includegraphics[width=0.91\columnwidth]{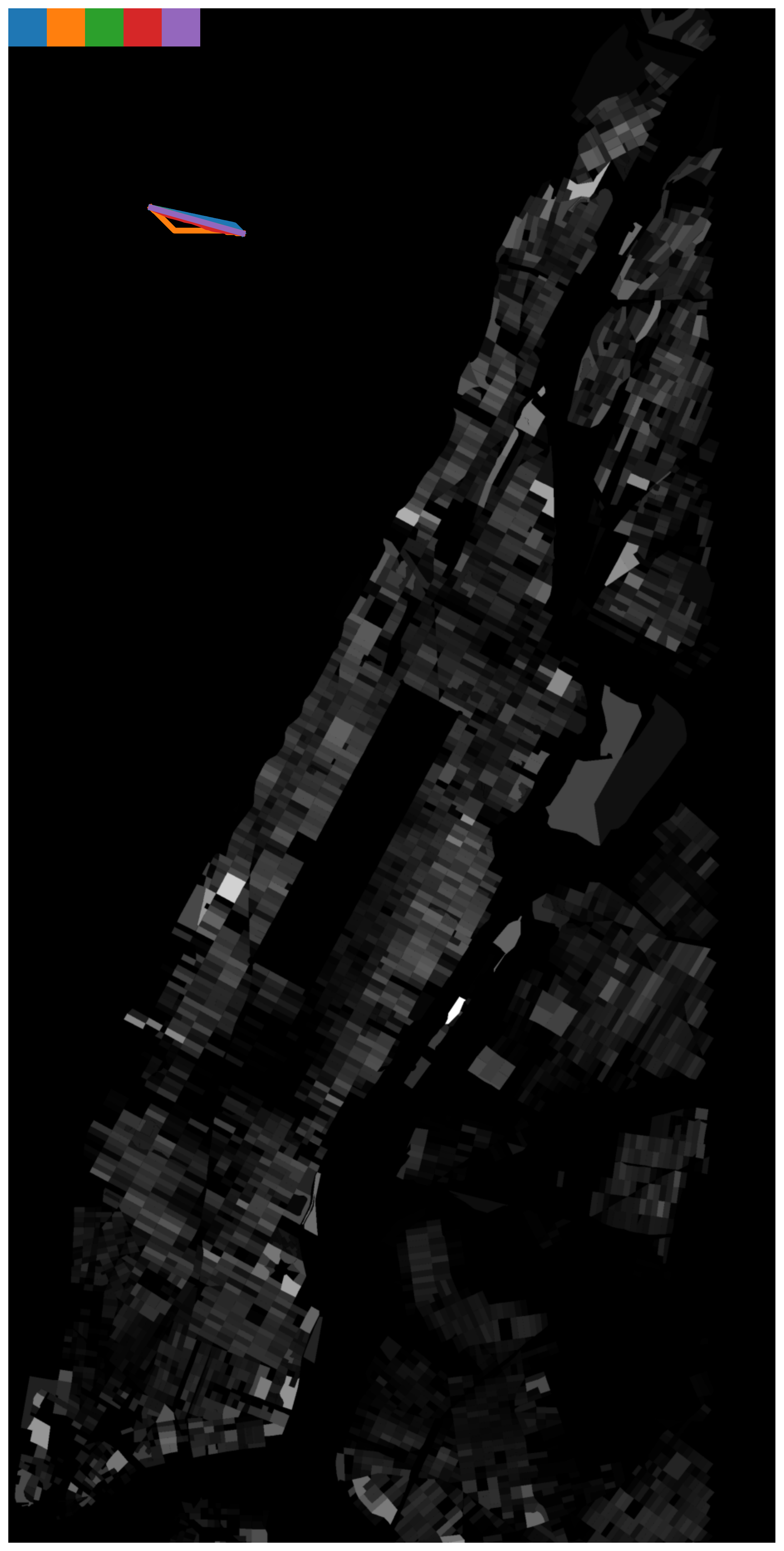}
        \caption{}
        \label{fig:ceil-alt-case-a}
    \end{subfigure}
    \hfill
    \begin{subfigure}{0.31\textwidth}
        \centering
        \includegraphics[width=0.91\columnwidth]{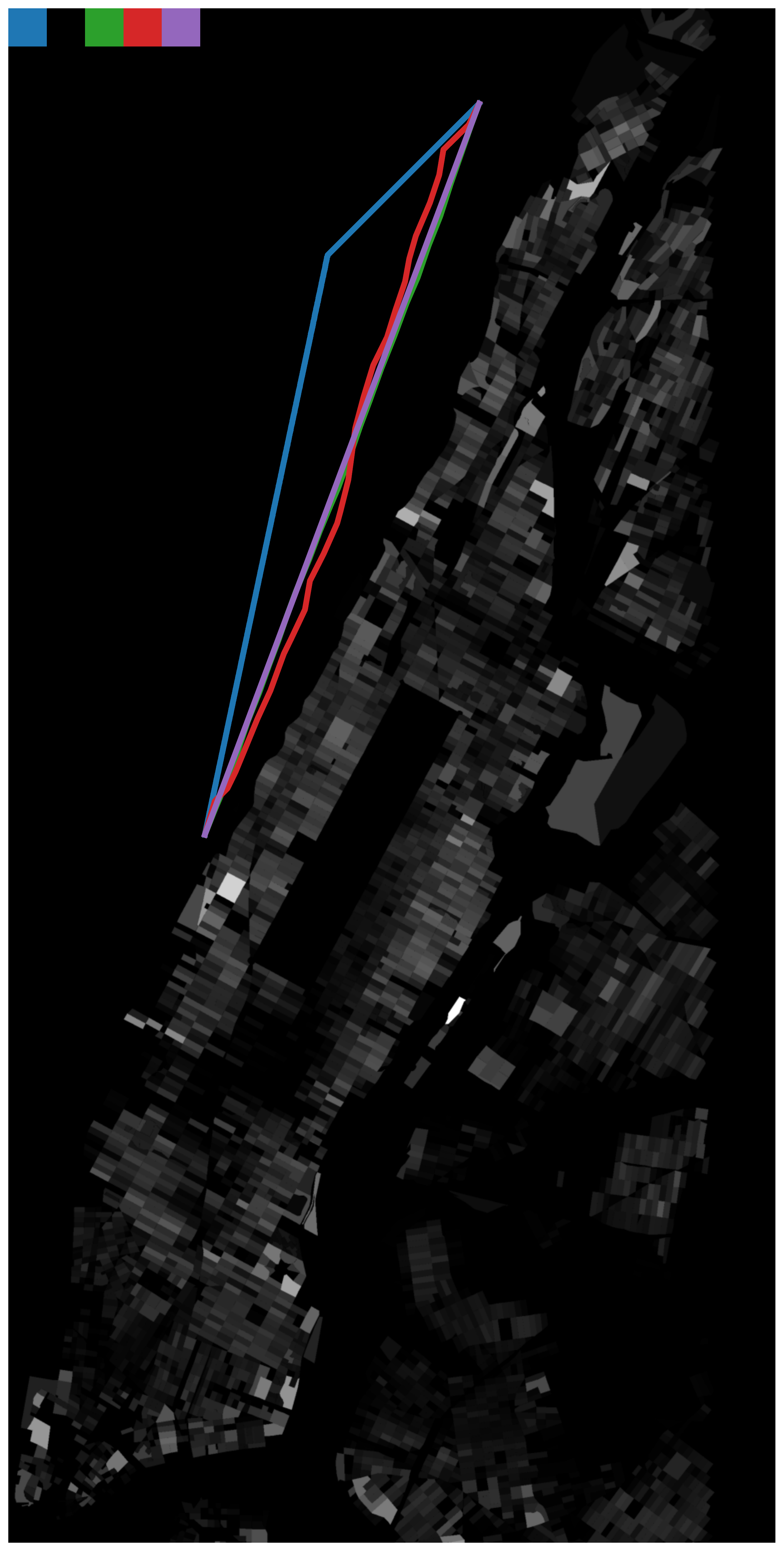}
        \caption{}
        \label{fig:ceil-alt-case-b}
    \end{subfigure}
    \hfill
    \begin{subfigure}{0.31\textwidth}
        \centering
        \includegraphics[width=0.91\columnwidth]{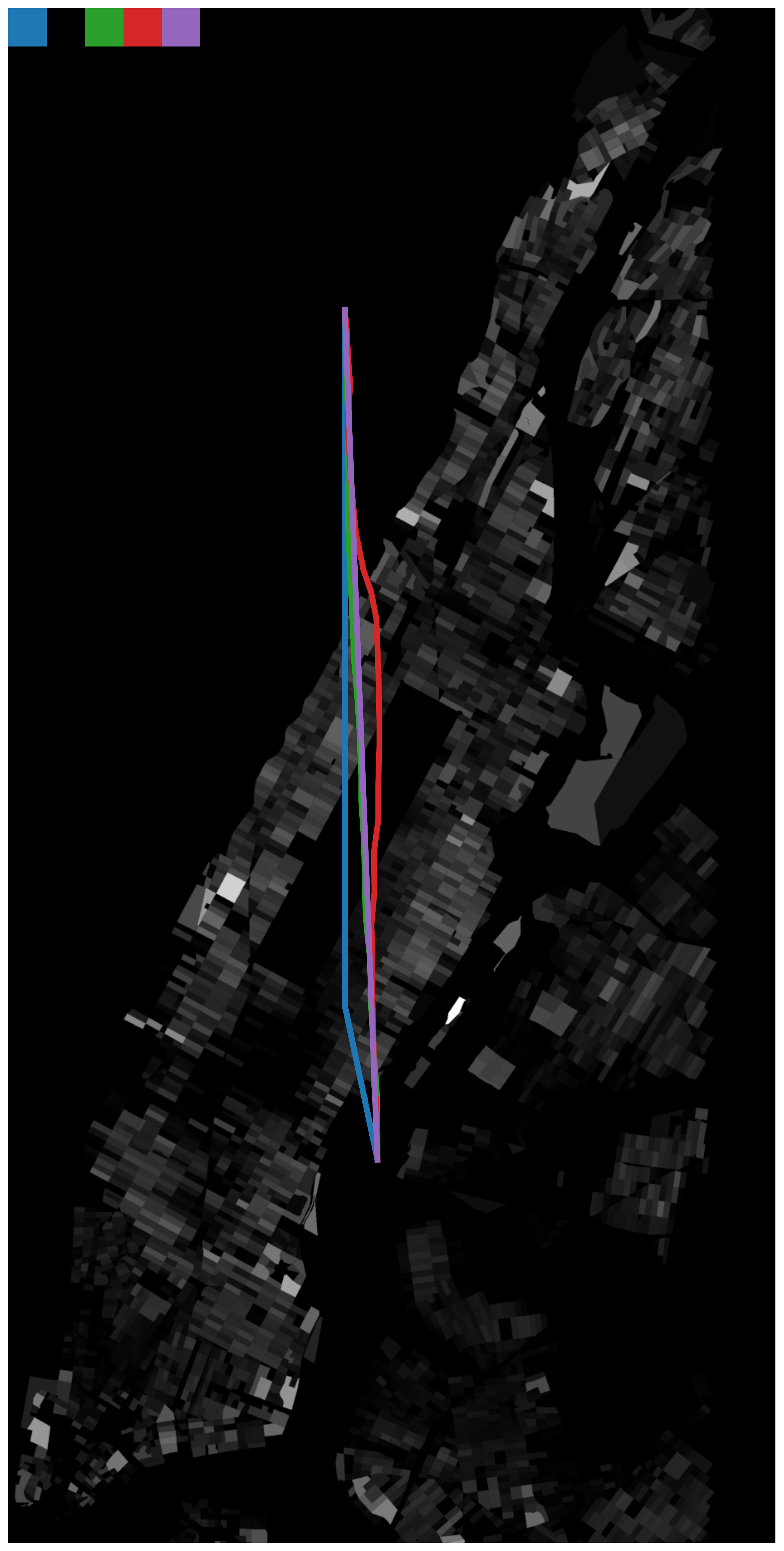}
        \caption{}
        \label{fig:ceil-alt-case-c}
    \end{subfigure}
    \begin{tabular}{c c c c c}
        \vspace{-09pt}\\
        \astar{dist} \textcolor{tab10-blue}{$\blacksquare$} & \astar{plus} \textcolor{tab10-orange}{$\blacksquare$} & \bitstar{dist} \textcolor{tab10-green}{$\blacksquare$} & \bitstar{plus} \textcolor{tab10-red}{$\blacksquare$} & \ptp{} \textcolor{tab10-purple}{$\blacksquare$} \\
        \vspace{-19pt}\\
    \end{tabular}
    \caption{Example solution paths at 600m AGL, 5m resolution maps in New York City.}
    \label{fig:ceil_alt_paths}
\end{figure}

Above all buildings at 600m AGL, only distance and population remain as nontrivial costs. As shown in Fig. \ref{fig:ceil-alt-case-a}, lack of obstacles and short travel distance is ideal for all planners. However, this may not be the case as range increases per Figs. \ref{fig:ceil-alt-case-c} and  \ref{fig:ceil-alt-case-c}. Along the Hudson River, distance is the only cost to optimize, making \ptp{} the best motion planner in this example. However, upon entering Manhattan, \bitstar{plus} becomes more suitable as it selects a route over lower population areas. The distance-population tradeoff demonstrates the benefits of geometric versus sampling-based planners. Collectively, these case studies illustrate the pros and cons of each planner thus motivate motion planning algorithm selection.

\section{Conclusion}
\label{sec:conclusions}

This paper has defined a set of map-based and path-based metrics for sUAS urban flight planning. Map-based metrics were investigated in detail with metric maps generated over Manhattan at three different resolutions for four sUAS AGL flight altitudes. Results demonstrate the complementary nature of GPS and lidar accuracy in an urban canyon as a function of altitude. By generating these metric maps a priori, an sUAS can predict risk and sensor data quality before a flight, i.e., GPS will provide valid position data if $m_{gps} > m_{lidar}$; lidar will offer better data otherwise. 

Population metric maps support residence-to-work and work-to-residence commuting patterns using as simplified as work-week daytime and nighttime models. In the future, this model should be extended to weekends with a time-based population function offering more resolution over 24-hour population patterns. When deep in the urban canyon, proximity-based risk is high, but it quickly decreases at higher altitudes due to fewer obstacles. For path planning, if risk is the primary cost, data indicate that flying to a higher altitude is preferable. Additional research is needed to incorporate risk metrics for urban flight planning, such as system, actuator, sensor, and weather-related risks, to extend current fixed-altitude maps to full 3D cost maps to support full 3D flight planning.

\bibliography{references}

\end{document}